\documentclass[sigconf, nonacm]{acmart}
\usepackage{diagbox}
\usepackage{microtype}
\usepackage{graphicx}
\usepackage{tabularx}
\usepackage{multirow}
\usepackage{makecell}
\usepackage{rotating}
\usepackage{booktabs}
\usepackage{subfigure}
\usepackage{booktabs}
\usepackage{amsmath}
\usepackage{mathtools}
\usepackage{amsthm}
\usepackage{balance}
\usepackage{hyperref}
\usepackage{xcolor}
\usepackage{algorithmic}
\usepackage[linesnumbered,ruled]{algorithm2e}
\newcommand\vldbdoi{XX.XX/XXX.XX}
\newcommand\vldbpages{XXX-XXX}
\newcommand\vldbvolume{14}
\newcommand\vldbissue{1}
\newcommand\vldbyear{2020}
\newcommand\vldbauthors{\authors}
\newcommand\vldbtitle{\shorttitle} 
\newcommand\vldbavailabilityurl{URL_TO_YOUR_ARTIFACTS}
\newcommand\vldbpagestyle{plain} 

\begin{document}
\title{How Much Can Time-related Features Enhance Time Series Forecasting?}

\author{Chaolv Zeng}
\affiliation{
    \institution{Shanghai Jiao Tong University}
    \city{Shanghai}
    \country{China}
}
\email{zclzcl@sjtu.edu.cn}

\author{Yuan Tian}
\affiliation{
    \institution{China Yangtze Power Co}
    \city{Yichang}
    \country{China}
}
\email{tian_yuan4@ctg.com.cn}

\author{Guanjie Zheng}
\authornote{Corresponding Author}
\affiliation{
    \institution{Shanghai Jiao Tong University}
    \city{Shanghai}
    \country{China}
}
\email{gjzheng@sjtu.edu.cn}

\author{Yunjun Gao}
\affiliation{
    \institution{Zhejiang University}
    \city{Hangzhou}
    \country{China}
}
\email{gaoyj@zju.edu.cn}

\begin{abstract}

Recent advancements in long-term time series forecasting (LTSF) have primarily focused on capturing cross-time and cross-variate (channel) dependencies within historical data.
However, a critical aspect often overlooked by many existing methods is the explicit incorporation of \textbf{time-related features} (e.g., season, month, day of the week, hour, minute), which are essential components of time series data.
The absence of this explicit time-related encoding limits the ability of current models to capture cyclical or seasonal trends and long-term dependencies, especially with limited historical input.
To address this gap, we introduce a simple yet highly efficient module designed to encode time-related features, Time Stamp Forecaster (TimeSter), thereby enhancing the backbone's forecasting performance.
By integrating TimeSter with a linear backbone, our model, TimeLinear, significantly improves the performance of a single linear projector, reducing MSE by an average of 23\% on benchmark datasets such as Electricity and Traffic.
Notably, TimeLinear achieves these gains while maintaining exceptional computational efficiency, delivering results that are on par with or exceed state-of-the-art models, despite using a fraction of the parameters.
\end{abstract}

\maketitle

\pagestyle{\vldbpagestyle}
\begingroup\small\noindent\raggedright\textbf{PVLDB Reference Format:}\\
\vldbauthors. \vldbtitle. PVLDB, \vldbvolume(\vldbissue): \vldbpages, \vldbyear.\\
\href{https://doi.org/\vldbdoi}{doi:\vldbdoi}
\endgroup
\begingroup
\renewcommand\thefootnote{}\footnote{\noindent
This work is licensed under the Creative Commons BY-NC-ND 4.0 International License. Visit \url{https://creativecommons.org/licenses/by-nc-nd/4.0/} to view a copy of this license. For any use beyond those covered by this license, obtain permission by emailing \href{mailto:info@vldb.org}{info@vldb.org}. Copyright is held by the owner/author(s). Publication rights licensed to the VLDB Endowment. \\
\raggedright Proceedings of the VLDB Endowment, Vol. \vldbvolume, No. \vldbissue\ %
ISSN 2150-8097. \\
\href{https://doi.org/\vldbdoi}{doi:\vldbdoi} \\
}\addtocounter{footnote}{-1}\endgroup

\ifdefempty{\vldbavailabilityurl}{}{
\vspace{.3cm}
\begingroup\small\noindent\raggedright\textbf{PVLDB Artifact Availability:}\\
The source code, data, and/or other artifacts have been made available at \url{https://github.com/zclzcl0223/TimeLinear}.
\endgroup
}

\section{Introduction}


Time series data~\cite{Bansal2021,khayati2020,paparrizos2022,SchmidlEtAl2022Anomaly,WenigEtAl2022TimeEval,zhong2024,qiu2024tfb} is a sequence of data collected at successive points in time, typically at uniform intervals.
Unlike other types of sequence data, such as text, time series data is characterized by each data point being associated with a time-related feature or time stamp\footnote{We use the terms “time-related feature” and “time stamp” interchangeably.} (e.g., season, month, week, hour, minute, etc), along with the corresponding value of the observed variable.
In cases where multiple variables are recorded, the time series transitions from univariate to multivariate series.
Given the ubiquity of multivariate time series data~\cite{bose2017}, accurate forecasting is essential across a broad range of fields, including transportation~\cite{lv2014traffic,fang2021,tran2020}, finance~\cite{xu2018stock,shi2024mambastock}, and climate~\cite{chen2024federatedpromptlearningweather,zhang2023skilful}.
In response to this demand, recent years have seen the development of numerous neural network architectures that have achieved substantial advancements in time series forecasting~\cite{zhou2021informer,cui2021,zeng2023transformers,wang2023timemixer,zhang2023crossformer,wu2021autoformer,Yuqietal-2023-PatchTST,liu2024itransformer,wu2023timesnet,luo2024moderntcn}.

\begin{figure}[tpb]
  \centering
  \includegraphics[width=1\columnwidth]{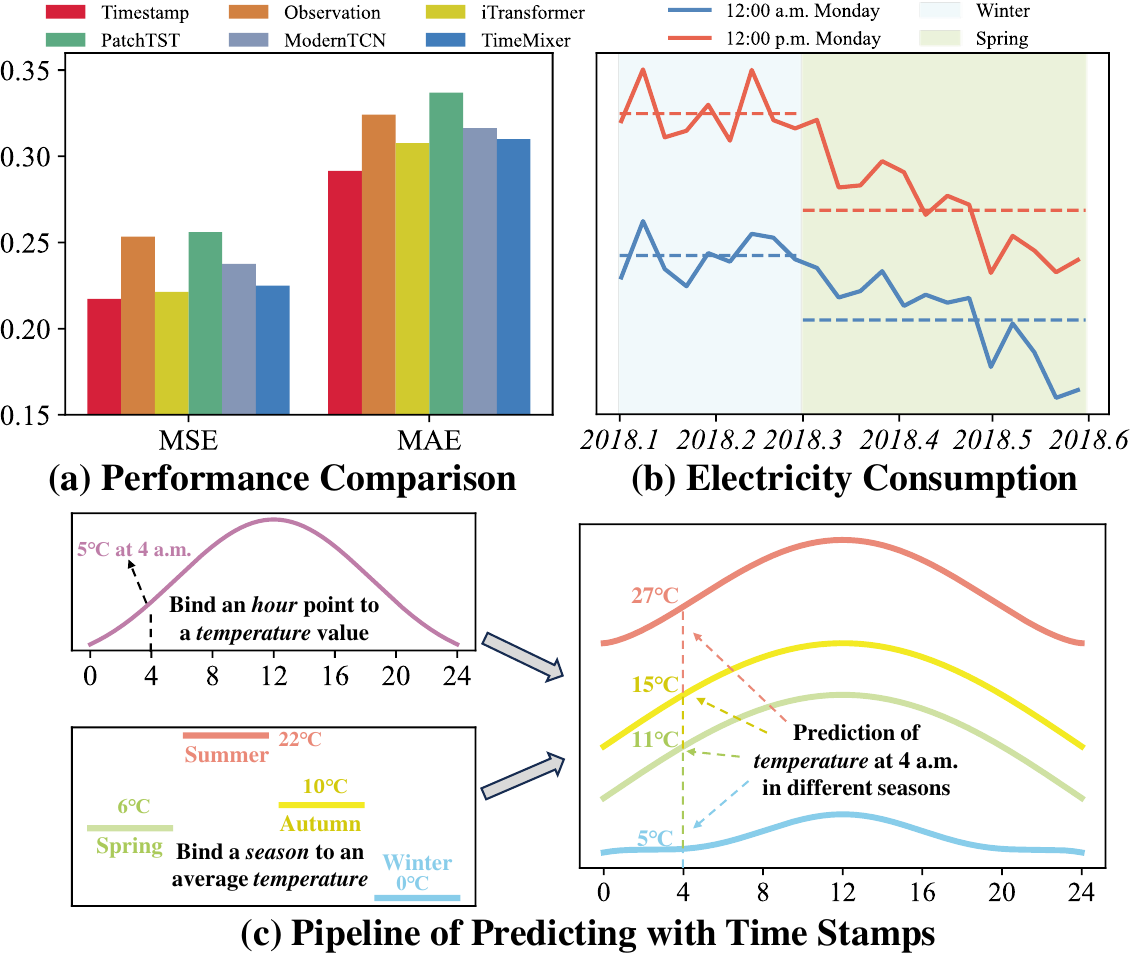}
  \caption{(a) Comparison of forecasting using only time-related features, historical observations, and other baselines on Electricity with a look-back window of 96 and a future horizon of 720.
  (b) 100 people's (indexed as \textit{190-290} on Electricity) average electricity consumption at noon and midnight every Monday during the first half of \textit{2018}.
  (c) Temperature prediction using time-related features, where each hour and season is associated with some temperature values.}
  \label{fig:intro}
\end{figure}

As mentioned above, each time series data point consists of a time stamp and the observed values of the measured variables.
However, most existing forecasting models primarily focus on modeling cross-time and cross-variate dependencies based on the historical observations of the variables, often overlooking the rich semantic information embedded in the associated time stamps.
Time stamps inherently provide crucial contextual information, as they constrain or dictate the values of the observed variables. 
For instance, in weather forecasting, summer months typically correlate with higher temperatures, while winter is associated with lower temperatures.
Similarly, in transportation, weekdays usually see heavier traffic, whereas weekends tend to have lower traffic volumes.
In other words, the combination of different time stamps corresponds to certain variable observations in a multidimensional vector space, within which variables fluctuate due to random noise.

To validate this claim, we conduct a case study using the Electricity dataset~\cite{wu2021autoformer},  where we encode the historical time stamps (hours, days, and seasons) to hidden space and make predictions based solely on this encoding using a single linear layer.
As illustrated in Figure~\ref{fig:intro} (a), where \textit{Timestamp} refers to models that use time-related features as input and \textit{Observation} refers to models that use historical observations of the variables as input, both approaches leverage a single linear layer for predictions.
Notably, under the setting of predicting the next 720 steps based on 96 historical data, forecasting based solely on time-related features outperforms all other baselines.
This can be attributed to the cyclical stability and seasonal patterns present in residential electricity consumption.
As depicted in Figure~\ref{fig:intro} (b), electricity consumption at noon and midnight on Mondays exhibits relatively stable behavior within the same season, especially in winter months, fluctuating around an average value, represented by the dotted line.
In contrast, values at the same time point differ significantly between spring and winter.
The association of these consumption patterns with specific time-related features highlights the potential of utilizing time stamps to enhance time series prediction.
Predicting with time stamps is analogous to binding a time-related feature to a specific observation, as shown in Figure~\ref{fig:intro} (c).
The combination of different time-related features produces different predictions accordingly.

Based on the above discussion and the widespread existence of the relationship between observations and time stamps across various types of time series (e.g., weather, traffic, energy), we propose to predict with time-related features.
However, given that we can only encode a limited number of time stamps (seasons, months, weeks, hours, minutes, etc.), while the future is unknown (such as years), relying solely on them would result in predictions that are invariant for all time points with the same time-related features.
Therefore, while time stamps provide a robust foundation, historical observations are still needed to introduce variability and cope with noise in the predictions.
Our proposed method consists of two key components to address multivariate time series forecasting with both time-related features and historical observations of variables.
The first part, named \textbf{Time Stamp Forecaster} (TimeSter), encodes historical time-related features into a hidden space and makes predictions with a single linear layer.
The second part, named \textbf{Backbone Forecaster} (BonSter), employs any other backbones (e.g., PatchTST~\cite{Yuqietal-2023-PatchTST}, iTransformer~\cite{liu2024itransformer}) to predict the noise components based on historical observations of variables.
The outputs of these two components are combined to yield the final prediction.
A simple combination of TimeSter and a linear layer backbone, named \textbf{TimeLinear}, can reduce the average MSE of a single linear projector on the Electricity and Traffic datasets by 23\%.
Technically, our contributions are summarized as:
\begin{itemize}
    \item Realizing the relationship between time stamps and time series observations, we comprehensively study the potential of time stamps in enhancing time series forecasting.
    \item We introduce a lightweight and effective module to leverage time-related features, Time Stamp Forecaster (\textbf{TimeSter}),  which can be easily integrated into various time series prediction models to boost their performance.
    \item  We present a simple yet powerful model, \textbf{TimeLinear}, the combination of TimeSter and a linear layer, which achieves consistent state-of-the-art performance across seven real-world datasets with only \textit{100k}$\sim$\textit{1M} parameters.
\end{itemize}

The rest of the paper will be organized as follows:
We first introduce the preliminary knowledge in Section~\ref{sec:pre} and related works in Section~\ref{sec:rela}.
Then we introduce our method in Section~\ref{sec:method}. In Section~\ref{sec:exp}, we present our experimental results and some visualization results in detail.
Finally, we conclude in Section~\ref{sec:conclu}.
\section{Preliminary}
\label{sec:pre}
\textbf{Multivariate Time Series Forecasting.}
Given historical observations of $V$ variables $\mathbf{X}=\{\mathbf{x}_1,...,\mathbf{x}_L\}\in\mathbb{R}^{L\times V}$, where $L$ denotes the look-back window, multivariate time series forecasting aims to predict the future values $\mathbf{Y}=\{\mathbf{x}_{L+1},...,\mathbf{x}_{L+T}\}\in\mathbb{R}^{T\times V}$, where $T$ is the prediction horizon.
Specifically, this work focuses on long-term time series forecasting, where the prediction length $T$ is greater than or equal to 96.
In the following sections, we denote multivariate series at a time point $i$ as $\mathbf{x}_i$ and univariate series as $x_i$.

\noindent\textbf{Time-related Features.}
Time-related features are auxiliary information that captures temporal information at the point when measurements are recorded.
These features typically include components such as season, month, day of the week, hour, and minute.
For instance, a time stamp like \textit{2024-07-15-14:10:00} can be represented by the categorical features $(Summer, July, Monday, 14h, 10min)$.
In practice, these categorical features are numerically encoded (e.g., $Summer= 1$, $July= 6$) and then normalized to a common range, such as $[-0.5, 0.5]$, to facilitate training in machine learning models.
In the following sections, we let $\mathbf{U}\in\mathbb{R}^{L\times r}$ denote the time-related features corresponding to the historical data, and $\mathbf{P}\in\mathbb{R}^{T\times r}$ denote the time-related features corresponding to the future data, where $r$ is the number of selected time-related features.
\section{Related Works}
\label{sec:rela}
\subsection{Predicting with Historical Observations}
The application of neural networks in time series forecasting has led to the development of numerous deep learning models tailored for this domain.
These models span a variety of architectures, including CNN-based models like MICN~\cite{wang2022micn}, TimesNet~\cite{wu2023timesnet}, and ModernTCN~\cite{luo2024moderntcn}; Transformer-based models such as Crossformer~\cite{zhang2023crossformer}, PatchTST~\cite{Yuqietal-2023-PatchTST}, and iTransformer~\cite{liu2024itransformer}; and models based on linear layers or MLPs, such as DLinear~\cite{zeng2023transformers}, RLinear~\cite{li2023revisiting}, FITS~\cite{xu2023fits}, TimeMixer~\cite{wang2023timemixer}, MSD-Mixer~\cite{zhong2024}, and SOFTS~\cite{han2024softs}.
Despite their architectural differences, these models share a common goal of enhancing the modeling of multivariate historical observations to improve forecasting accuracy.
Some models, such as DLinear, RLinear, FITS, PatchTST, and TimeMixer, focus on capturing long-range temporal dependencies, while others, like Crossformer, ModernTCN, TimesNet, iTransformer, and SOFTS, emphasize modeling the interactions among variables.
However, a common limitation across these models is the lack of explicit consideration for the relationship between time-related features and the corresponding observations, which reduces their sensitivity to temporal dynamics, such as seasonal patterns and cyclical trends.

\subsection{Predicting with Time-related Features}
Incorporating time-related features, such as seasonality, day of the week, and time of day, is essential to improve predictive accuracy in time series forecasting.
Models like Autoformer~\cite{wu2021autoformer} and TimesNet~\cite{wu2023timesnet} simply add time stamps to historical series through position encoding.
iTransformer~\cite{liu2024itransformer} treats different time-related features as different tokens and feeds them to the attention layer together with other time series variables.
However, these models often treat time-related features as secondary, leading to suboptimal modeling of the interactions between these features and the target variables.
This is further evidenced by empirical results~\cite{wang2024rethinking}, which demonstrate that removing time-related features from these models has minimal impact on performance.
Many other models simply ignore time-related information~\cite{zeng2023transformers,Yuqietal-2023-PatchTST,xu2023fits,luo2024moderntcn}.
With the advent of LLMs and time series foundation models~\cite{goswami2024moment,ansari2024chronoslearninglanguagetime}, recent work~\cite{jin2023time,liu2024autotimes,narasimhan2024time,wang2024news} has explored time-related features as prompts or metadata, yielding promising results.
In smaller models, efforts have emerged to explicitly model time-related features and capture periodicity.
GLAFF~\cite{wang2024rethinking} applies a transformer to model dependencies between time-related features across different time points and then generates adaptive weights to balance global and local information.
However, GLAFF lacks precise alignment between time-related features and historical observations, and its computational demands are high.
More importantly, these methods that utilize time-related features all ignore the selection of time stamps, which is proven to be crucial in our experiments.
CycleNet~\cite{lin2024cyclenet} learns a periodic sequence for each time series to model repetition-based periodicity.
Nonetheless, it focuses solely on static periodic patterns and overlooks dynamic, timestamp-driven variations such as seasonality.

\section{Method}
\label{sec:method}
\begin{figure}[tpb]
  \centering
  \includegraphics[width=1\columnwidth]{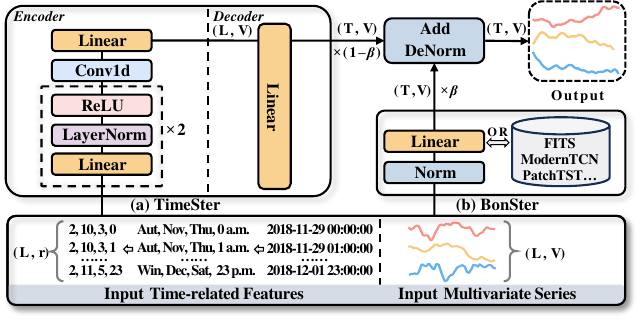}
  \caption{Overview of the proposed method.
  (a) The TimeSter module encodes historical time-related features and predicts future values.
  (b) The BonSter, i.e., any backbone model, utilizes historical observations of multivariate time series to generate predictions.
  Their outputs are weighted and summed to yield the final prediction.
  The model after \textit{Norm} could be replaced with any backbone.
  Here, we use a linear layer and name the whole model \textbf{TimeLinear}.}
  \label{fig:model}
\end{figure}

In this section, we present TimeSter, our proposed module that makes predictions based on time-related features, and describe how it integrates with various backbone models.
An overview of the architecture is shown in Figure~\ref{fig:model}.
To evaluate the effectiveness of our approach, we consider the linear layer backbone and the resulting model, \textbf{TimeLinear}, as our main model.
\subsection{Time Stamp Forecaster}
\begin{figure}[htpb]
  \centering
  \includegraphics[width=1\columnwidth]{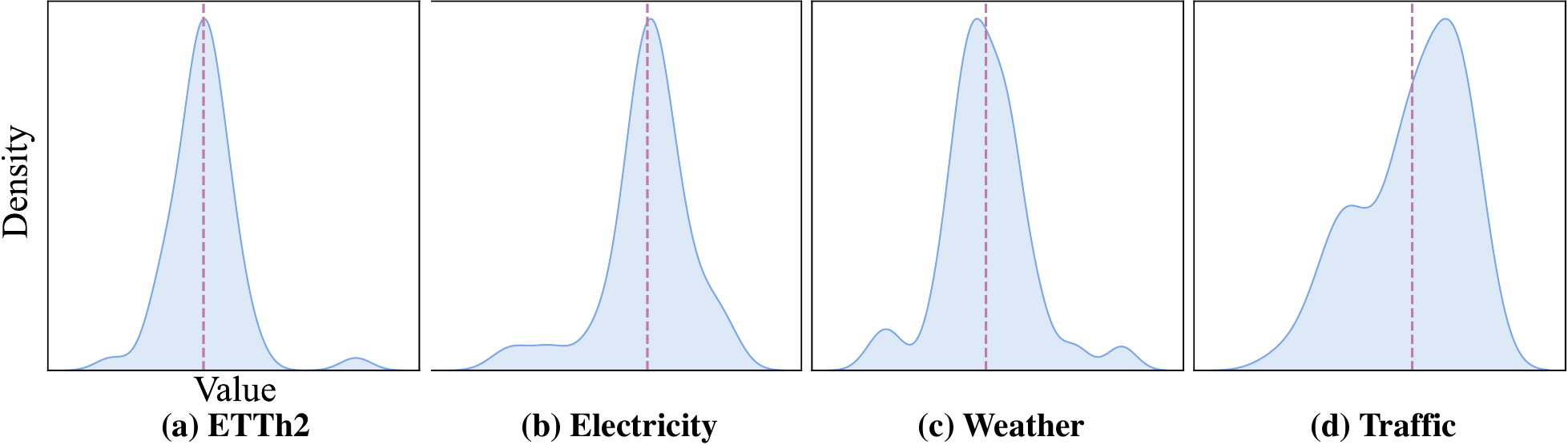}
  \caption{Distribution of training data across datasets at noon on Mondays. From left to right, ETTh2: Variate \textit{HUFL}, Electricity: Variate \textit{1}, Weather: Variate \textit{p (mbar)}, and Traffic: Variate \textit{1}. From this, we can see a significant timestamp-related distribution.}
  \label{fig:distribution}
\end{figure}

\textbf{Time Stamp Encoder.}
Given time-related features $\mathbf{U}_i\in\mathbb{R}^{r}$ at a specific time point $i$, where $r$ denotes the number of time-related features, the observation of a variable at this time $x_i$ is assumed to follow the conditional distribution $p_\phi(x_i|\mathbf{U}_i)$.
As depicted in Figure~\ref{fig:distribution}, we visualize the probability density distribution of a selected training variable across different datasets at noon every Monday.
The results indicate a time-dependent distribution pattern for these variables, yet directly modeling this distribution remains challenging.
Inspired by~\cite{kingma2013auto}, we approximate this distribution with a learnable model $q_\theta$.
For generality, we consider the distribution of $V$ variables over an $L$-step historical window:
\begin{equation}
\begin{aligned}
\mathbf{X}_\mathbf{U}\sim q_\theta(\mathbf{X}_\mathbf{U}|\mathbf{U})\,,
\end{aligned}
    \label{formula:1}
\end{equation}
where $\mathbf{U}\in\mathbb{R}^{L\times r}$ represents historical time-related features and $\mathbf{X}_\mathbf{U}\in\mathbb{R}^{L\times V}$ represents the generated time-related observations.
As shown in Figure~\ref{fig:model} (a) left, we treat $q_\theta$ as an encoder consisting of two nonlinear hidden layers, a one-dimensional convolution layer, and a linear projection layer.
Each linear layer projects along time-related feature and multivariate observation dimensions.
The convolutional layer fuses features in the same dimension as the above linear layers and mixes channels in the time dimension to generate diverse outputs.

\noindent\textbf{Time-related Observation Decoder.}
After generating the time-related multivariate observations, a single linear layer predicts the future observations, as depicted in Figure~\ref{fig:model} (a) right:
\begin{equation}
\begin{aligned}
\mathbf{Y}_\mathbf{U}= \mathbf{W}\mathbf{X}_\mathbf{U}+\mathbf{b}\,,
\end{aligned}
    \label{formula:2}
\end{equation}
where $\mathbf{W}\in\mathbb{R}^{T\times L}$ and $\mathbf{b}\in\mathbb{R}^{T}$ are learnable parameters of the linear layer.
The output $\mathbf{Y}_\mathbf{U}\in\mathbb{R}^{T\times V}$ represents future predictions based on historical time-related observations.

\subsection{Backbone Forecaster}
Historical observations of time series are fundamental in time series modeling, as they directly capture the changing patterns of time series and provide a basis for accurate forecasting.
To leverage both historical observations and time-related features, we incorporate other backbone models to process historical observations, complementing the TimeSter module.
This integration allows TimeSter to be easily incorporated with various backbones.
Technically, for the historical multivariate observations $\mathbf{X}\in\mathbb{R}^{L\times V}$, predictions are made with a specified backbone model, such as PatchTST~\cite{Yuqietal-2023-PatchTST}:
\begin{equation}
\begin{aligned}
\mathbf{Y}_\mathbf{B}= \text{Backbone}(\mathbf{X})\,,
\end{aligned}
    \label{formula:3}
\end{equation}
where $\mathbf{Y}_\mathbf{B}\in\mathbb{R}^{T\times V}$ represents predictions based on historical observations.
To isolate the effects of the backbone model and highlight the effectiveness of our time-related feature modeling, we use a single linear layer as our backbone in the following sections:
\begin{equation}
\begin{aligned}
\mathbf{Y}_\mathbf{B}= \mathbf{W}_\mathbf{B}\mathbf{X}+\mathbf{b}_\mathbf{B}\,.
\end{aligned}
    \label{formula:5}
\end{equation}
Combining TimeSter, this model variant is referred to as \textbf{TimeLinear}.
Finally, both predictions $\mathbf{Y}_\mathbf{U}$ and $\mathbf{Y}_\mathbf{B}$ are weighted and combined to produce the final result:
\begin{equation}
\begin{aligned}
\mathbf{Y}'= \mathbf{\beta}\mathbf{Y}_\mathbf{B}+(1-\mathbf{\beta})\mathbf{Y}_\mathbf{U}\,,
\end{aligned}
    \label{formula:4}
\end{equation}
where $\mathbf{Y}'\in\mathbb{R}^{T\times V}$ is the prediction, $\mathbf{\beta}\in(0,1)$ is a fixed coefficient that balances the contributions from each module.
The full process is depicted in Figure~\ref{fig:model} (b).

\subsection{Simplified Reversible Instance Normalization}
The Reversible Instance Normalization (RevIN)~\cite{kim2021reversible} is a widely used technique to mitigate distribution shifts in time series data.
Since it was proposed, it has been widely used in almost all    models~\cite{li2023revisiting,xu2023fits,wu2023timesnet,Yuqietal-2023-PatchTST,liu2024itransformer,luo2024moderntcn,han2024softs,lin2024cyclenet} in recent years and has played a significant role in improving their performance.
To maintain consistent normalization across backbone models, we implement RevIN in all cases.
For TimeLinear, we use a simplified RevIN variant without learnable affine parameters, as recent studies~\cite{liu2024itransformer,lin2024cyclenet} indicate it offers comparable performance with reduced parameters.
Technically, we normalize historical observations with their respective means and standard deviations, and then denormalize the final forecasts using these same statistics.
Note that we denormalize the final forecasts, not the output of the backbone network.
Similar denormalization strategies also apply to other backbone networks.
The process can be formulated as:
\begin{equation}
\begin{aligned}
\mathbf{\hat{X}}&= \frac{\mathbf{X}-\mathbf{\mu}}{\sqrt{\mathbf{\sigma}^2+\mathbf{\epsilon}}}\,,\\
\mathbf{\hat{Y}}'&= \mathbf{Y}'\times\sqrt{\mathbf{\sigma}^2+\mathbf{\epsilon}}+\mathbf{\mu}\,,
\end{aligned}
    \label{formula:6}
\end{equation}
where $\mathbf{\mu}$, $\mathbf{\sigma}\in\mathbb{R}^V$ are the means and standard deviations of the multivariate historical observations, and $\mathbf{\epsilon}$ is a small constant to avoid division by zero.
The entire prediction process of TimeLinear is shown in Algorithm~\ref{alg1}, where the \textit{Linear} layer could be replaced with any other backbone models.

\begin{algorithm}[tpb]
\caption{Pipeline of TimeLinear}
\renewcommand{\algorithmicrequire}{\textbf{Input:}}
\renewcommand{\algorithmicensure}{\textbf{Output:}}
\label{alg1}
\begin{algorithmic}[1]
\REQUIRE Historical time series $\mathbf{X}\in\mathbb{R}^{L\times V}$; historical time stamp feature $\mathbf{U}\in\mathbb{R}^{L\times r}$; future horizon $T$; trade-off coefficient $\beta$
\STATE $\mu,\sigma\gets\text{Mean}(\mathbf{X}),\text{STD}(\mathbf{X})$\quad// {$\mu,\sigma\in\mathbb{R}^V$}
\STATE $\mathbf{\hat{X}}\gets\frac{\mathbf{X}-\mathbf{\mu}}{\sqrt{\mathbf{\sigma}^2+\mathbf{\epsilon}}}$\space\space\space\space\space\space\quad\quad\quad\quad\quad// {$\mathbf{\hat{X}}\in\mathbb{R}^{L\times V}$}
\STATE $\mathbf{Y}_\mathbf{U}\gets\text{TimeSter}(\mathbf{U})$\space\space\quad\quad\quad// {$\mathbf{Y}_\mathbf{U}\in\mathbb{R}^{T\times V}$}
\STATE $\mathbf{Y}_\mathbf{B}\gets\text{Linear}(\mathbf{\hat{X}})$\space\space\quad\quad\quad\quad// {$\mathbf{Y}_\mathbf{B}\in\mathbb{R}^{T\times V}$}
\STATE $\mathbf{Y}'\gets\mathbf{\beta}\mathbf{Y}_\mathbf{B}+(1-\mathbf{\beta})\mathbf{Y}_\mathbf{U}$\quad\quad// {$\mathbf{Y}'\in\mathbb{R}^{T\times V}$}
\STATE $\mathbf{\hat{Y}}'\gets\mathbf{Y}'\times\sqrt{\mathbf{\sigma}^2+\mathbf{\epsilon}}+\mathbf{\mu}$\quad\quad// {$\mathbf{\hat{Y}}'\in\mathbb{R}^{T\times V}$}
\ENSURE $\mathbf{\hat{Y}}'$\quad\quad\quad\quad\quad\quad\quad\quad// The final prediction
\end{algorithmic}
\end{algorithm}

\section{Experiments}
\label{sec:exp}
In this section, we conduct a series of experiments to comprehensively evaluate the validity and effectiveness of our proposed method.
We begin by detailing the experimental settings.
In Section~\ref{sec:mainresults}, we present the performance improvements achieved by the explicit incorporation of time-related features.
Subsequently, in Section~\ref{sec:ablationstudies}, we perform ablation studies to systematically verify the soundness of our modeling approach for time-related features.
Finally, in Sections~\ref{sec:moreanalysis} and~\ref{sec:visualization}, we provide an in-depth analysis of why time-related features enhance performance, supported by additional insights and visualizations.

\begin{table}[htpb]
\caption{Overview of dataset. $Variable$ denotes the number of variables. $Timespan$ indicates the duration. $Granularity$ indicates the interval between two time steps. $Domain$ denotes the physical meaning of the observed value.}
\centering
\label{tab:dataset}
\resizebox{1.0\linewidth}{!}{
\begin{tabular}{c|c|c|c|c}
\toprule[1pt]
Dataset & Variable &Timespan& Granularity & Domain \\
\toprule[1pt]
ETTm1 \& ETTm2 & 7 & 2016.7 - 2018.6 & 15 minutes & Electricity \\

ETTh1 \& ETTh2 & 7 & 2016.7 - 2018.6 & 1 hour & Electricity \\

Electricity & 321 & 2016.7 - 2019.7 & 1 hour & Electricity \\

Weather & 21 & 2020.1 - 2021.1 & 10 minutes & Weather \\

Traffic & 862 & 2016.7 - 2018.7 & 1 hour & Transportation \\

\bottomrule[1pt]
\end{tabular}
}
\end{table}
\noindent\textbf{Datasets.}
We evaluate TimeLinear on seven well-built real-world datasets: four ETT series (ETTm1, ETTm2, ETTh1, ETTh2), Electricity, Weather, and Traffic, as detailed in Table~\ref{tab:dataset}.
We adopt the dataset split and normalization approach used in previous studies~\cite{wu2021autoformer,wu2023timesnet}.

\noindent\textbf{Baselines.}
We compare TimeLinear with models spanning various architectures:
(i) Linear-based: DLinear~\cite{zeng2023transformers}, FITS~\cite{xu2023fits}, RLinear~\cite{li2023revisiting}, CycleNet~\cite{lin2024cyclenet};
(ii) MLP-based: TimeMixer~\cite{wang2023timemixer}, TiDE~\cite{das2023long}, MSD-Mixer~\cite{zhong2024}, SOFTS~\cite{han2024softs};
(iii) Convolution-based: TimesNet~\cite{wu2023timesnet}, ModernTCN~\cite{luo2024moderntcn};
and (iv) Transformer-based: Crossformer~\cite{zhang2023crossformer}, PatchTST~\cite{Yuqietal-2023-PatchTST}, iTransformer~\cite{liu2024itransformer}.

Linear-based models, including our TimeLinear backbone RLinear (consisting of a linear layer and the simplified RevIN), use a single linear layer for predictions, hence, they are simpler than MLP-based models that employ multi-layer perceptrons for encoding.
Moreover, GLAFF~\cite{wang2024rethinking}, which generates adaptive weights from time-related features to balance global and local dependencies, is included as a linear baseline with the same backbone as TimeLinear, denoted as GLAFFLinear.
\begin{table*}[htpb]
\caption{Results of the multivariate long-term forecasting task under prediction lengths $T\in\{96,192,336,720\}$. 
The historical window $L$ is fixed at $96$.
We report the average performance of four prediction lengths.
For Linear-based models, we highlight the best in \textcolor{red}{red} and the runner-up in \underline{\textcolor{blue}{blue}}.
For all architectures, we highlight the best performers with *.
}
\centering
\label{tab:main}
\resizebox{0.99\linewidth}{!}{
\begin{tabular}{p{0.2cm}|c|cc|cc|cc|cc|cc|cc|cc}

\toprule[1pt]
\multicolumn{2}{c}{Dataset} &
\multicolumn{2}{c}{ETTm1}  & \multicolumn{2}{c}{ETTm2} & \multicolumn{2}{c}{ETTh1} & \multicolumn{2}{c}{ETTh2} & \multicolumn{2}{c}{Electricity} & \multicolumn{2}{c}{Weather} & \multicolumn{2}{c}{Traffic}  \\
\cmidrule(l{10pt}r{10pt}){1-2}\cmidrule(l{10pt}r{10pt}){3-4}\cmidrule(l{10pt}r{10pt}){5-6}\cmidrule(l{10pt}r{10pt}){7-8}\cmidrule(l{10pt}r{10pt}){9-10}\cmidrule(l{10pt}r{10pt}){11-12}\cmidrule(l{10pt}r{10pt}){13-14}\cmidrule(l{10pt}r{10pt}){15-16}
\multicolumn{2}{c}{Metric} & MSE & MAE & MSE & MAE & MSE & MAE & MSE & MAE & MSE & MAE & MSE & MAE & MSE & MAE \\
\toprule[1pt]

\multirow{9}{*}{\rotatebox[origin=c]{90}{Others}} & TimesNet~\cite{wu2023timesnet} & 0.400 & 0.406 & 0.291 & 0.333 & 0.458 & 0.450 & 0.414 & 0.427 & 0.192 & 0.295 & 0.259 & 0.287 & 0.620 & 0.336 \\

& TiDE~\cite{das2023long} & 0.419 & 0.419 & 0.358 & 0.404 & 0.541 & 0.507 & 0.611 & 0.550 & 0.251 & 0.344 & 0.271 & 0.320 & 0.760 & 0.473 \\

& Crossformer~\cite{zhang2023crossformer} & 0.513 & 0.496 & 0.757 & 0.610 & 0.529 & 0.522 & 0.942 & 0.684 & 0.244 & 0.334 & 0.259 & 0.315 & 0.550 & 0.304 \\

& PatchTST~\cite{Yuqietal-2023-PatchTST} & 0.387 & 0.400 & 0.283 & 0.327 & 0.445 & 0.441 & 0.378 & 0.403 & 0.191 & 0.280 & 0.256 & 0.278 & 0.461 & 0.291 \\

& MSD-Mixer~\cite{zhong2024} & 0.361* & 0.387* & 0.271* & 0.324 & 0.440 & 0.433 & 0.368 & 0.406 & 0.173 & 0.272 & 0.233* & 0.287 & 0.524 & 0.340 \\

& iTransformer~\cite{liu2024itransformer} & 0.407 & 0.410 & 0.288 & 0.332 & 0.454 & 0.447 & 0.383 & 0.407 & 0.178 & 0.270 & 0.258 & 0.278 & 0.428 & 0.282 \\

& TimeMixer~\cite{wang2023timemixer} & 0.381 & 0.395 & 0.275 & 0.323 & 0.447 & 0.440 & 0.364 & 0.395 & 0.182 & 0.272 & 0.240 & 0.271* & 0.484 & 0.297 \\

& ModernTCN~\cite{luo2024moderntcn} & 0.386 & 0.401 & 0.278 & 0.322 & 0.445 & 0.432 & 0.381 & 0.404 & 0.197 & 0.282 & 0.240 & 0.271* & 0.550 & 0.369 \\

& SOFTS~\cite{han2024softs} & 0.393 & 0.403 & 0.287 & 0.330 & 0.449 & 0.442 & 0.373 & 0.400 & 0.174 & 0.264 & 0.255 & 0.278 & 0.409* & 0.267* \\

\midrule

\multirow{6}{*}{\rotatebox[origin=c]{90}{Linear}} & DLinear~\cite{zeng2023transformers} & 0.403 & 0.407 & 0.350 & 0.401 & 0.456 & 0.452 & 0.559 & 0.515 & 0.212 & 0.300 & 0.265 & 0.317 & 0.625 & 0.383 \\

& RLinear~\cite{li2023revisiting} & 0.412 & \underline{\textcolor{blue}{0.406}} & 0.286 & \underline{\textcolor{blue}{0.327}} & 0.446 & 0.433 & 0.377 & 0.399 & 0.215 & 0.291 & 0.273 & 0.291 & 0.623 & 0.371 \\

& FITS~\cite{xu2023fits} & 0.415 & 0.408 & 0.286 & 0.328 & 0.444 & 0.432 & \underline{\textcolor{blue}{0.374}} & \underline{\textcolor{blue}{0.397}} & 0.217 & 0.295 & 0.273 & 0.292 & 0.627 & 0.375 \\

& CycleNet~\cite{lin2024cyclenet} & \underline{\textcolor{blue}{0.386}} & \textcolor{red}{0.395} & \textcolor{red}{0.272} & \textcolor{red}{0.315}* & \underline{\textcolor{blue}{0.433}} & \underline{\textcolor{blue}{0.428}} & 0.384 & 0.405 & \underline{\textcolor{blue}{0.170}} & \underline{\textcolor{blue}{0.261}} & \underline{\textcolor{blue}{0.255}} & \underline{\textcolor{blue}{0.279}} & \underline{\textcolor{blue}{0.486}} & 0.313 \\

& GLAFFLinear~\cite{wang2024rethinking} & 0.422 & 0.416 & 0.289 & \underline{\textcolor{blue}{0.327}} & 0.459 & 0.444 & 0.394 & 0.411 & 0.204 & 0.297 & 0.273 & 0.292 & 0.562 & \textcolor{red}{0.301} \\

\cmidrule(l{10pt}r{10pt}){2-16}

& TimeLinear [Ours] & \textcolor{red}{0.385} & \textcolor{red}{0.395} & \underline{\textcolor{blue}{0.273}} & \textcolor{red}{0.315}* & \textcolor{red}{0.432}* & \textcolor{red}{0.426}* & \textcolor{red}{0.358}* & \textcolor{red}{0.389}* & \textcolor{red}{0.165}* & \textcolor{red}{0.259}* & \textcolor{red}{0.251} & \textcolor{red}{0.276} & \textcolor{red}{0.480} & \underline{\textcolor{blue}{0.304}} \\

\bottomrule[1pt]
\end{tabular}
}
\end{table*}

\noindent\textbf{Implementation.}
All models use a historical window length of $L=96$ and are trained using Adam~\cite{kingma2014adam} with MSE loss.
We evaluate models with Mean Squared Error (MSE) and Mean Absolute Error (MAE) over four prediction horizons $T\in\{96,192,336,720\}$.
Each experiment is repeated three times, and we report the mean values.
All codes are implemented in Pytorch~\cite{paszke2019pytorch}.
We conduct experiments on a single NVIDIA GeForce RTX 3090 24G GPU.

\subsection{Main Results}
In the main experiment, we start by comparing TimeLinear with state-of-the-art baseline models to validate the effectiveness of incorporating time-related features.
Next, we integrate the TimeSter with various backbones to demonstrate the general applicability of our proposed method.
\label{sec:mainresults}

\subsubsection{Long-term Forecasting Performance}
Table~\ref{tab:main} presents the multivariate long-term forecasting results across TimeLinear and other baselines.
Most baseline results follow the original papers, except when experimental settings differ, in which case we rerun them (e.g., ModernTCN, PatchTST) under the official hyperparameters.
The results show that TimeLinear achieves leading performance.
On the one hand, TimeLinear is the best-performing Linear-based model, ranking top 1 in \textbf{12} out of 14 settings among all Linear-based models. 
On the other hand, TimeLinear is comparable to those models with complex encoders, ranking top 1 in \textbf{7} out of 14 settings among all methods.
These outstanding performances are not only seen in datasets with a small number of variables, such as the ETT dataset (7 variables) but also in datasets with a large number of variables, such as the Electricity dataset (321 variables), where the complex correlations between variables have a great impact on the prediction results.
\begin{figure}[htpb]
  \centering
  \includegraphics[width=1\columnwidth]{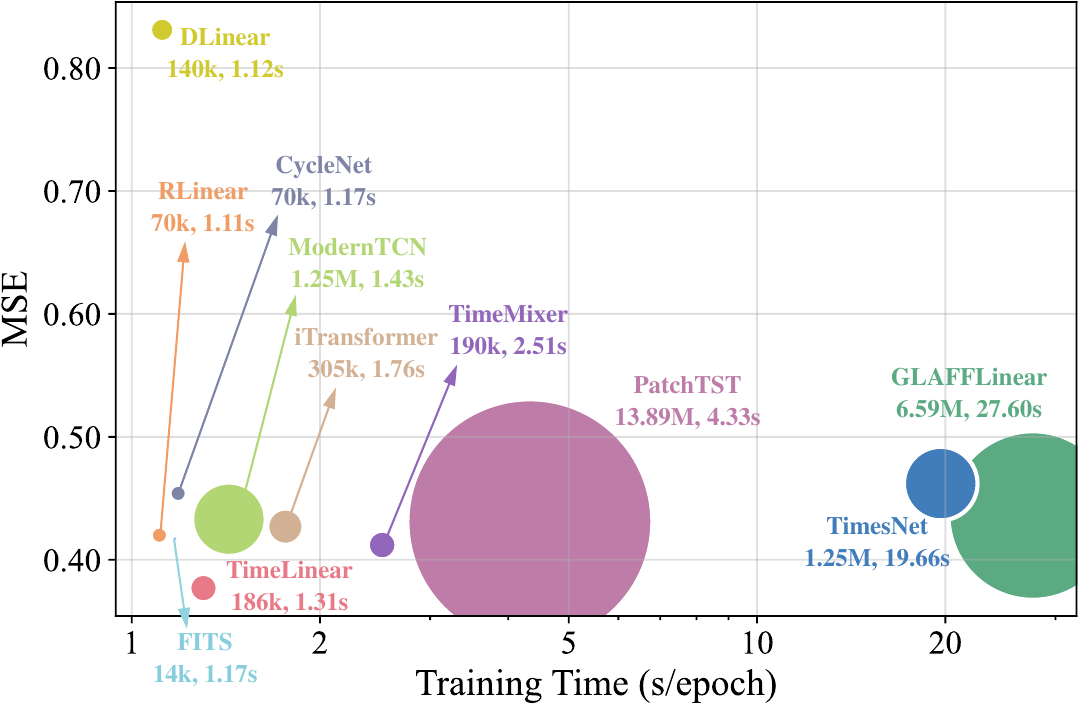}
  \caption{Accuracy, parameter count, and training time of TimeLinear and other models on ETTh2, with historical window $L=96$ and future horizon $T=720$.}
  \label{fig:eff}
\end{figure}
More remarkably, TimeLinear achieves these with significantly fewer parameters and faster training speed than all non-linear models.
Figure~\ref{fig:eff} shows that TimeLinear, with only \textit{100k} parameters,  outperforms other models and demonstrates superior efficiency.
The results fully demonstrate the importance of time-related features and the effectiveness of our modeling.

\subsubsection{Generalization Performance}
To further verify the versatility of TimeSter, we evaluate its integration with various backbones.
In addition to RLinear (the backbone of TimeLinear), we analyze TimeSter on five more models across different architectures and domains: FITS (Linear-based model in the frequency domain), ModernTCN and TimesNet (Convolution-based), iTransformer and PatchTST (Transformer-based).
For a fair comparison, all results are reproduced on the same machine under the optimal hyperparameters, so some results might be different from Table~\ref{tab:main}.
\begin{table*}[htpb]
\caption{Performance gain after combining TimeSter with channel-independent backbones, i.e., RLinear (Linear-based), FITS (Linear-based), and PatchTST (Transformer-based).
The look-back window is $L=96$.
The performance of four prediction lengths $T\in\{96,192,336,720\}$ is reported.
\textit{Avg} indicates the average performance of four prediction lengths.
\textcolor{red}{Red} denotes improved performance and \textcolor[HTML]{5CAA82}{Green} indicates decreasing performance.
\textit{Dataset Avg} denotes the average improvement of a model on all datasets.}
\centering
\label{tab:generalization_full_ci}
\resizebox{0.98\linewidth}{!}{
\begin{tabular}{p{0.2cm}c|cc|cc|cc|cc|cc|cc|cc|cc|cc}

\toprule[1pt]

\multicolumn{2}{c|}{Models} & \multicolumn{2}{c}{\makecell{RLinear~\cite{li2023revisiting}}} & \multicolumn{2}{c}{w / TimeSter} & \multicolumn{2}{c|}{Improvement} & \multicolumn{2}{c}{\makecell{FITS~\cite{xu2023fits}}} & \multicolumn{2}{c}{w / TimeSter} & \multicolumn{2}{c}{Improvement} & \multicolumn{2}{c}{\makecell{PatchTST~\cite{Yuqietal-2023-PatchTST}}} & \multicolumn{2}{c}{w / TimeSter} & \multicolumn{2}{c}{Improvement} \\

\cmidrule(l{10pt}r{10pt}){3-4}\cmidrule(l{10pt}r{10pt}){5-6}\cmidrule(l{10pt}r{10pt}){7-8}\cmidrule(l{10pt}r{10pt}){9-10}\cmidrule(l{10pt}r{10pt}){11-12}\cmidrule(l{10pt}r{10pt}){13-14}\cmidrule(l{10pt}r{10pt}){15-16}\cmidrule(l{10pt}r{10pt}){17-18}\cmidrule(l{10pt}r{10pt}){19-20}

\multicolumn{2}{c|}{Metric} & MSE & MAE  & MSE & MAE & MSE & MAE & MSE & MAE & MSE & MAE & MSE & MAE & MSE & MAE & MSE & MAE & MSE & MAE \\

\toprule[1pt]
\multirow{4}{*}{\rotatebox[origin=c]{90}{ETTm1}}
& \multicolumn{1}{|c|}{96} & 0.350 & 0.370 & 0.325 & 0.364 & \textcolor{red}{7.06\%} & \textcolor{red}{1.78\%} & 0.353 & 0.374 & 0.329 & 0.368 & \textcolor{red}{6.77\%} & \textcolor{red}{1.73\%} & 0.325  & 0.364  & 0.316  & 0.363  & \textcolor{red}{3.00\%} & \textcolor{red}{0.36\%} \\

& \multicolumn{1}{|c|}{192} & 0.389 & 0.390 & 0.365 & 0.381 & \textcolor{red}{6.13\%} & \textcolor{red}{2.26\%} & 0.392 & 0.393 & 0.366 & 0.383 & \textcolor{red}{6.55\%} & \textcolor{red}{2.49\%} & 0.368  & 0.388  & 0.362  & 0.390  & \textcolor{red}{1.67\%} & \textcolor[HTML]{5CAA82}{-0.67\%} \\

& \multicolumn{1}{|c|}{336} & 0.422 & 0.413 & 0.395 & 0.401 & \textcolor{red}{6.43\%} & \textcolor{red}{2.88\%} & 0.425 & 0.415 & 0.397 & 0.402 & \textcolor{red}{6.67\%} & \textcolor{red}{2.98\%} & 0.396  & 0.405  & 0.387  & 0.405  & \textcolor{red}{2.42\%} & \textcolor[HTML]{5CAA82}{-0.03\%} \\

& \multicolumn{1}{|c|}{720} & 0.486 & 0.448 & 0.456 & 0.433 & \textcolor{red}{6.23\%} & \textcolor{red}{3.47\%} & 0.488 & 0.450 & 0.457 & 0.434 & \textcolor{red}{6.42\%} & \textcolor{red}{3.57\%} & 0.459  & 0.444  & 0.451  & 0.446  & \textcolor{red}{1.72\%} & \textcolor[HTML]{5CAA82}{-0.34\%} \\

\cmidrule(l{10pt}r{10pt}){2-20}
& \multicolumn{1}{|c|}{Avg} & 0.412 & 0.406 & 0.385 & 0.395 & \textcolor{red}{6.43\%} & \textcolor{red}{2.64\%} & 0.415 & 0.408 & 0.387 & 0.397 & \textcolor{red}{6.59\%} & \textcolor{red}{2.74\%} & 0.387  & 0.400  & 0.379  & 0.401  & \textcolor{red}{2.15\%} & \textcolor[HTML]{5CAA82}{-0.18\%} \\

\midrule
\multirow{4}{*}{\rotatebox[origin=c]{90}{ETTm2}}
& \multicolumn{1}{|c|}{96}& 0.183 & 0.264 & 0.167 & 0.249 & \textcolor{red}{8.48\%} & \textcolor{red}{5.97\%} & 0.183 & 0.266 & 0.168 & 0.249 & \textcolor{red}{8.51\%} & \textcolor{red}{6.25\%} & 0.179  & 0.260  & 0.170  & 0.254  & \textcolor{red}{4.96\%} & \textcolor{red}{2.28\%} \\

& \multicolumn{1}{|c|}{192} & 0.246 & 0.304 & 0.233 & 0.291 & \textcolor{red}{5.41\%} & \textcolor{red}{4.28\%} & 0.247 & 0.305 & 0.234 & 0.292 & \textcolor{red}{5.45\%} & \textcolor{red}{4.44\%} & 0.242  & 0.302  & 0.241  & 0.302  & \textcolor{red}{0.45\%} & \textcolor{red}{0.00\%} \\

& \multicolumn{1}{|c|}{336} & 0.307 & 0.342 & 0.295 & 0.331 & \textcolor{red}{4.11\%} & \textcolor{red}{3.24\%} & 0.308 & 0.343 & 0.295 & 0.332 & \textcolor{red}{4.14\%} & \textcolor{red}{3.34\%} & 0.306  & 0.345  & 0.298  & 0.340  & \textcolor{red}{2.50\%} & \textcolor{red}{1.43\%} \\

& \multicolumn{1}{|c|}{720} & 0.407 & 0.398 & 0.395 & 0.390 & \textcolor{red}{2.96\%} & \textcolor{red}{2.03\%} & 0.408 & 0.398 & 0.396 & 0.390 & \textcolor{red}{2.92\%} & \textcolor{red}{2.25\%} & 0.406  & 0.402  & 0.401  & 0.400  & \textcolor{red}{1.30\%} & \textcolor{red}{0.39\%} \\

\cmidrule(l{10pt}r{10pt}){2-20}
& \multicolumn{1}{|c|}{Avg} & 0.286 & 0.327 & 0.273 & 0.315 & \textcolor{red}{4.68\%} & \textcolor{red}{3.67\%} & 0.286 & 0.328 & 0.273 & 0.316 & \textcolor{red}{4.69\%} & \textcolor{red}{3.85\%} & 0.283  & 0.327  & 0.278  & 0.324  & \textcolor{red}{2.02\%} & \textcolor{red}{0.95\%} \\

\midrule
\multirow{4}{*}{\rotatebox[origin=c]{90}{ETTh1}}
& \multicolumn{1}{|c|}{96}& 0.389 & 0.395 & 0.378 & 0.391 & \textcolor{red}{2.75\%} & \textcolor{red}{1.00\%} & 0.386 & 0.394 & 0.378 & 0.392 & \textcolor{red}{2.00\%} & \textcolor{red}{0.42\%} & 0.377  & 0.396  & 0.377  & 0.396  & \textcolor{red}{0.03\%} & \textcolor{red}{0.10\%} \\

& \multicolumn{1}{|c|}{192} & 0.437 & 0.424 & 0.424 & 0.418 & \textcolor{red}{2.95\%} & \textcolor{red}{1.24\%} & 0.438 & 0.424 & 0.424 & 0.418 & \textcolor{red}{3.34\%} & \textcolor{red}{1.48\%} & 0.425  & 0.427  & 0.422  & 0.425  & \textcolor{red}{0.74\%} & \textcolor{red}{0.38\%} \\

& \multicolumn{1}{|c|}{336} & 0.479 & 0.445 & 0.463 & 0.438 & \textcolor{red}{3.29\%} & \textcolor{red}{1.41\%} & 0.479 & 0.444 & 0.460 & 0.436 & \textcolor{red}{3.89\%} & \textcolor{red}{1.85\%} & 0.461  & 0.447  & 0.459  & 0.446  & \textcolor{red}{0.52\%} & \textcolor{red}{0.31\%} \\

& \multicolumn{1}{|c|}{720} & 0.480 & 0.469 & 0.464 & 0.456 & \textcolor{red}{3.35\%} & \textcolor{red}{2.75\%} & 0.474 & 0.464 & 0.464 & 0.456 & \textcolor{red}{1.92\%} & \textcolor{red}{1.80\%} & 0.518  & 0.493  & 0.496  & 0.489  & \textcolor{red}{4.29\%} & \textcolor{red}{0.89\%} \\

\cmidrule(l{10pt}r{10pt}){2-20}
& \multicolumn{1}{|c|}{Avg} & 0.446 & 0.433 & 0.432 & 0.426 & \textcolor{red}{3.11\%} & \textcolor{red}{1.64\%} & 0.444 & 0.432 & 0.432 & 0.425 & \textcolor{red}{2.82\%} & \textcolor{red}{1.42\%} & 0.445  & 0.441  & 0.438  & 0.439  & \textcolor{red}{1.57\%} & \textcolor{red}{0.44\%} \\

\midrule
\multirow{4}{*}{\rotatebox[origin=c]{90}{ETTh2}}
& \multicolumn{1}{|c|}{96}& 0.291 & 0.339 & 0.285 & 0.335 & \textcolor{red}{1.93\%} & \textcolor{red}{1.17\%} & 0.290 & 0.339 & 0.287 & 0.337 & \textcolor{red}{0.97\%} & \textcolor{red}{0.53\%} & 0.294  & 0.343  & 0.290  & 0.340  & \textcolor{red}{1.51\%} & \textcolor{red}{0.73\%} \\

& \multicolumn{1}{|c|}{192} & 0.376 & 0.390 & 0.373 & 0.390 & \textcolor{red}{0.80\%} & \textcolor{red}{0.07\%} & 0.375 & 0.390 & 0.373 & 0.392 & \textcolor{red}{0.42\%} & \textcolor[HTML]{5CAA82}{-0.49\%} & 0.377  & 0.398  & 0.364  & 0.389  & \textcolor{red}{3.41\%} & \textcolor{red}{2.14\%} \\

& \multicolumn{1}{|c|}{336} & 0.420 & 0.427 & 0.398 & 0.418 & \textcolor{red}{5.16\%} & \textcolor{red}{2.26\%} & 0.414 & 0.424 & 0.405 & 0.423 & \textcolor{red}{2.10\%} & \textcolor{red}{0.26\%} & 0.414  & 0.426  & 0.408  & 0.426  & \textcolor{red}{1.39\%} & \textcolor{red}{0.01\%} \\

& \multicolumn{1}{|c|}{720} & 0.422 & 0.440 & 0.377 & 0.412 & \textcolor{red}{10.78\%} & \textcolor{red}{6.41\%} & 0.417 & 0.436 & 0.386 & 0.418 & \textcolor{red}{7.43\%} & \textcolor{red}{4.18\%} & 0.426  & 0.445  & 0.421  & 0.441  & \textcolor{red}{1.15\%} & \textcolor{red}{0.89\%} \\

\cmidrule(l{10pt}r{10pt}){2-20}
& \multicolumn{1}{|c|}{Avg} & 0.377 & 0.399 & 0.358 & 0.389 & \textcolor{red}{5.02\%} & \textcolor{red}{2.64\%} & 0.374 & 0.397 & 0.363 & 0.393 & \textcolor{red}{2.94\%} & \textcolor{red}{1.21\%} & 0.378  & 0.403  & 0.371  & 0.399  & \textcolor{red}{1.85\%} & \textcolor{red}{0.93\%} \\

\midrule
\multirow{4}{*}{\rotatebox[origin=c]{90}{Electricity}}
& \multicolumn{1}{|c|}{96} & 0.197 & 0.274 & 0.140 & 0.234 & \textcolor{red}{29.10\%} & \textcolor{red}{14.48\%} & 0.200 & 0.278 & 0.141 & 0.237 & \textcolor{red}{29.28\%} & \textcolor{red}{14.83\%} & 0.165  & 0.256  & 0.142  & 0.240  & \textcolor{red}{13.92\%} & \textcolor{red}{5.95\%} \\

& \multicolumn{1}{|c|}{192} & 0.197 & 0.276 & 0.155 & 0.247 & \textcolor{red}{21.57\%} & \textcolor{red}{10.46\%} & 0.199 & 0.280 & 0.156 & 0.250 & \textcolor{red}{21.78\%} & \textcolor{red}{10.82\%} & 0.174  & 0.265  & 0.160  & 0.258  & \textcolor{red}{8.33\%} & \textcolor{red}{2.69\%} \\

& \multicolumn{1}{|c|}{336} & 0.212 & 0.292 & 0.169 & 0.265 & \textcolor{red}{20.10\%} & \textcolor{red}{9.11\%} & 0.214 & 0.295 & 0.171 & 0.268 & \textcolor{red}{20.18\%} & \textcolor{red}{9.34\%} & 0.191  & 0.282  & 0.173  & 0.273  & \textcolor{red}{9.50\%} & \textcolor{red}{3.13\%} \\

& \multicolumn{1}{|c|}{720} & 0.253 & 0.324 & 0.198 & 0.290 & \textcolor{red}{21.91\%} & \textcolor{red}{10.43\%} & 0.255 & 0.327 & 0.200 & 0.293 & \textcolor{red}{21.69\%} & \textcolor{red}{10.37\%} & 0.232  & 0.316  & 0.202  & 0.300  & \textcolor{red}{12.84\%} & \textcolor{red}{5.17\%} \\

\cmidrule(l{10pt}r{10pt}){2-20}
& \multicolumn{1}{|c|}{Avg} & 0.215 & 0.291 & 0.165 & 0.259 & \textcolor{red}{23.04\%} & \textcolor{red}{11.06\%} & 0.217 & 0.295 & 0.167 & 0.262 & \textcolor{red}{23.09\%} & \textcolor{red}{11.27\%} & 0.191  & 0.280  & 0.169  & 0.268  & \textcolor{red}{11.20\%} & \textcolor{red}{4.25\%} \\

\midrule
\multirow{4}{*}{\rotatebox[origin=c]{90}{Weather}}
& \multicolumn{1}{|c|}{96} & 0.195 & 0.234 & 0.166 & 0.212 & \textcolor{red}{14.91\%} & \textcolor{red}{9.29\%} & 0.194 & 0.235 & 0.167 & 0.213 & \textcolor{red}{14.38\%} & \textcolor{red}{9.16\%} & 0.173  & 0.214  & 0.157  & 0.205  & \textcolor{red}{9.49\%} & \textcolor{red}{4.27\%} \\

& \multicolumn{1}{|c|}{192} & 0.240 & 0.270 & 0.218 & 0.256 & \textcolor{red}{9.03\%} & \textcolor{red}{5.37\%} & 0.240 & 0.271 & 0.219 & 0.256 & \textcolor{red}{8.92\%} & \textcolor{red}{5.50\%} & 0.219  & 0.255  & 0.208  & 0.250  & \textcolor{red}{5.12\%} & \textcolor{red}{2.31\%} \\

& \multicolumn{1}{|c|}{336} & 0.291 & 0.306 & 0.272 & 0.294 & \textcolor{red}{6.60\%} & \textcolor{red}{4.02\%} & 0.292 & 0.307 & 0.273 & 0.294 & \textcolor{red}{6.59\%} & \textcolor{red}{4.21\%} & 0.277  & 0.297  & 0.265  & 0.290  & \textcolor{red}{4.58\%} & \textcolor{red}{2.22\%} \\

& \multicolumn{1}{|c|}{720} & 0.364 & 0.353 & 0.347 & 0.342 & \textcolor{red}{4.81\%} & \textcolor{red}{3.05\%}  & 0.365 & 0.354 & 0.347 & 0.342 & \textcolor{red}{4.85\%} & \textcolor{red}{3.19\%} & 0.354  & 0.347  & 0.342  & 0.341  & \textcolor{red}{3.30\%} & \textcolor{red}{1.54\%} \\

\cmidrule(l{10pt}r{10pt}){2-20}
& \multicolumn{1}{|c|}{Avg} & 0.273 & 0.291 & 0.251 & 0.276 & \textcolor{red}{8.02\%} & \textcolor{red}{5.10\%} & 0.273 & 0.292 & 0.251 & 0.277 & \textcolor{red}{7.91\%} & \textcolor{red}{5.20\%} & 0.256  & 0.278  & 0.243  & 0.272  & \textcolor{red}{5.08\%} & \textcolor{red}{2.42\%} \\

\midrule
\multirow{4}{*}{\rotatebox[origin=c]{90}{Traffic}}
& \multicolumn{1}{|c|}{96}& 0.645 & 0.383 & 0.459 & 0.293 & \textcolor{red}{28.83\%} & \textcolor{red}{23.45\%} & 0.650 & 0.387 & 0.468 & 0.307 & \textcolor{red}{27.96\%} & \textcolor{red}{20.59\%} & 0.437  & 0.280  & 0.416  & 0.279  & \textcolor{red}{4.79\%} & \textcolor{red}{0.08\%} \\

& \multicolumn{1}{|c|}{192} & 0.598 & 0.359 & 0.467 & 0.298 & \textcolor{red}{21.83\%} & \textcolor{red}{16.92\%} & 0.602 & 0.363 & 0.480 & 0.315 & \textcolor{red}{20.17\%} & \textcolor{red}{13.28\%} & 0.448  & 0.284  & 0.429  & 0.284  & \textcolor{red}{4.24\%} & \textcolor[HTML]{5CAA82}{-0.23\%} \\

& \multicolumn{1}{|c|}{336} & 0.605 & 0.362 & 0.481 & 0.305 & \textcolor{red}{20.50\%} & \textcolor{red}{15.60\%} & 0.609 & 0.365 & 0.488 & 0.313 & \textcolor{red}{19.82\%} & \textcolor{red}{14.27\%} & 0.463  & 0.292  & 0.444  & 0.291  & \textcolor{red}{4.15\%} & \textcolor{red}{0.05\%} \\

& \multicolumn{1}{|c|}{720} & 0.643 & 0.381 & 0.512 & 0.320 & \textcolor{red}{20.43\%} & \textcolor{red}{16.11\%} & 0.647 & 0.385 & 0.516 & 0.328 & \textcolor{red}{20.23\%} & \textcolor{red}{14.90\%} & 0.497  & 0.310  & 0.477  & 0.309  & \textcolor{red}{3.91\%} & \textcolor{red}{0.17\%} \\

\cmidrule(l{10pt}r{10pt}){2-20}
& \multicolumn{1}{|c|}{Avg} & 0.623 & 0.371 & 0.480 & 0.304 & \textcolor{red}{22.96\%} & \textcolor{red}{18.07\%} & 0.627 & 0.375 & 0.488 & 0.316 & \textcolor{red}{22.12\%} & \textcolor{red}{15.82\%} & 0.461  & 0.291  & 0.442  & 0.291  & \textcolor{red}{4.26\%} & \textcolor{red}{0.02\%} \\

\midrule
\multicolumn{2}{c|}{Dataset Avg} & \multicolumn{4}{c|}{} & \textcolor{red}{10.47\%} & \textcolor{red}{6.40\%} & \multicolumn{4}{c|}{} & \textcolor{red}{10.02\%} & \textcolor{red}{5.93\%} & \multicolumn{4}{c|}{} & \textcolor{red}{4.02\%} & \textcolor{red}{1.26\%} \\

\bottomrule[1pt]
\end{tabular}
}
\end{table*}

\begin{table*}[htpb]
\caption{Performance gain after combining TimeSter with channel-dependent backbones, i.e., ModernTCN (Convolution-based), TimesNet (Convolution-based), and iTransformer (Transformer-based).
The look-back window is $L=96$.
The performance of four prediction lengths $T\in\{96,192,336,720\}$ is reported.
\textit{Avg} indicates the average performance of four prediction lengths.
\textcolor{red}{Red} denotes improved performance and \textcolor[HTML]{5CAA82}{Green} indicates decreasing performance.
\textit{Dataset Avg} denotes the average improvement of a model on all datasets.}
\centering
\label{tab:generalization_full_cd}
\resizebox{0.98\linewidth}{!}{
\begin{tabular}{p{0.2cm}c|cc|cc|cc|cc|cc|cc|cc|cc|cc}

\toprule[1pt]

\multicolumn{2}{c|}{Models} & \multicolumn{2}{c}{\makecell{ModernTCN~\cite{luo2024moderntcn}}} & \multicolumn{2}{c}{w / TimeSter} & \multicolumn{2}{c|}{Improvement} & \multicolumn{2}{c}{\makecell{TimesNet~\cite{wu2023timesnet}}} & \multicolumn{2}{c}{w / TimeSter} & \multicolumn{2}{c}{Improvement} & \multicolumn{2}{c}{\makecell{iTransformer~\cite{liu2024itransformer}}} & \multicolumn{2}{c}{w / TimeSter} & \multicolumn{2}{c}{Improvement} \\

\cmidrule(l{10pt}r{10pt}){3-4}\cmidrule(l{10pt}r{10pt}){5-6}\cmidrule(l{10pt}r{10pt}){7-8}\cmidrule(l{10pt}r{10pt}){9-10}\cmidrule(l{10pt}r{10pt}){11-12}\cmidrule(l{10pt}r{10pt}){13-14}\cmidrule(l{10pt}r{10pt}){15-16}\cmidrule(l{10pt}r{10pt}){17-18}\cmidrule(l{10pt}r{10pt}){19-20}

\multicolumn{2}{c|}{Metric} & MSE & MAE  & MSE & MAE & MSE & MAE & MSE & MAE & MSE & MAE & MSE & MAE & MSE & MAE & MSE & MAE & MSE & MAE \\

\toprule[1pt]
\multirow{4}{*}{\rotatebox[origin=c]{90}{ETTm1}}
& \multicolumn{1}{|c|}{96} & 0.317 & 0.362 & 0.315 & 0.359 & \textcolor{red}{0.59\%} & \textcolor{red}{0.64\%} & 0.331  & 0.371  & 0.334  & 0.371 & \textcolor[HTML]{5CAA82}{-1.01\%} & \textcolor[HTML]{5CAA82}{-0.05\%} & 0.341  & 0.376  & 0.325  & 0.369  & \textcolor{red}{4.60\%} & \textcolor{red}{1.85\%} \\

& \multicolumn{1}{|c|}{192} & 0.363 & 0.389 & 0.358 & 0.387 & \textcolor{red}{1.62\%} & \textcolor{red}{0.55\%} & 0.392  & 0.404  & 0.379  & 0.396 & \textcolor{red}{3.40\%} & \textcolor{red}{1.75\%} & 0.382  & 0.395  & 0.367  & 0.390  & \textcolor{red}{3.90\%} & \textcolor{red}{1.47\%} \\

& \multicolumn{1}{|c|}{336} & 0.403 & 0.412 & 0.399 & 0.411 & \textcolor{red}{1.03\%} & \textcolor{red}{0.06\%} & 0.418  & 0.424  & 0.413  & 0.422 & \textcolor{red}{1.05\%} & \textcolor{red}{0.51\%} & 0.419  & 0.419  & 0.402  & 0.411  & \textcolor{red}{4.07\%} & \textcolor{red}{1.78\%} \\

& \multicolumn{1}{|c|}{720} & 0.461 & 0.443 & 0.457 & 0.439 & \textcolor{red}{0.94\%} & \textcolor{red}{0.96\%} & 0.490  & 0.460  & 0.480  & 0.456 & \textcolor{red}{1.90\%} & \textcolor{red}{1.05\%}  & 0.490  & 0.458  & 0.469  & 0.448  & \textcolor{red}{4.39\%} & \textcolor{red}{2.31\%} \\

\cmidrule(l{10pt}r{10pt}){2-20}
& \multicolumn{1}{|c|}{Avg} & 0.386 & 0.401 & 0.382 & 0.399 & \textcolor{red}{1.05\%} & \textcolor{red}{0.56\%} & 0.408  & 0.415  & 0.402  & 0.411 & \textcolor{red}{1.45\%} & \textcolor{red}{0.84\%}  & 0.408  & 0.412  & 0.391  & 0.404  & \textcolor{red}{4.24\%} & \textcolor{red}{1.87\%} \\

\midrule
\multirow{4}{*}{\rotatebox[origin=c]{90}{ETTm2}}
& \multicolumn{1}{|c|}{96} & 0.173 & 0.255 & 0.169 & 0.252 & \textcolor{red}{1.85\%} & \textcolor{red}{1.14\%} & 0.188  & 0.266  & 0.186  & 0.264 & \textcolor{red}{0.82\%} & \textcolor{red}{0.66\%} & 0.185  & 0.270  & 0.174  & 0.258  & \textcolor{red}{5.80\%} & \textcolor{red}{4.64\%} \\

& \multicolumn{1}{|c|}{192} & 0.235 & 0.296 & 0.237 & 0.297 & \textcolor[HTML]{5CAA82}{-0.88\%} & \textcolor[HTML]{5CAA82}{-0.20\%} & 0.255  & 0.309  & 0.256  & 0.308 & \textcolor[HTML]{5CAA82}{-0.65\%} & \textcolor{red}{0.43\%} & 0.252  & 0.312  & 0.243  & 0.303  & \textcolor{red}{3.55\%} & \textcolor{red}{3.06\%} \\

& \multicolumn{1}{|c|}{336} & 0.308 & 0.344 & 0.299 & 0.336 & \textcolor{red}{2.80\%} & \textcolor{red}{2.20\%} & 0.319  & 0.347  & 0.312  & 0.343 & \textcolor{red}{2.23\%} & \textcolor{red}{1.12\%} & 0.314  & 0.351  & 0.302  & 0.340  & \textcolor{red}{3.79\%} & \textcolor{red}{3.07\%} \\

& \multicolumn{1}{|c|}{720} & 0.398 & 0.394 & 0.395 & 0.392 & \textcolor{red}{0.61\%} & \textcolor{red}{0.34\%} & 0.419  & 0.406  & 0.413  & 0.401 & \textcolor{red}{1.39\%} & \textcolor{red}{1.05\%} & 0.412  & 0.406  & 0.403  & 0.398  & \textcolor{red}{2.23\%} & \textcolor{red}{2.00\%} \\

\cmidrule(l{10pt}r{10pt}){2-20}
& \multicolumn{1}{|c|}{Avg} & 0.278 & 0.322 & 0.275 & 0.319 & \textcolor{red}{1.09\%} & \textcolor{red}{0.87\%} & 0.295  & 0.332  & 0.292  & 0.329 & \textcolor{red}{1.08\%} & \textcolor{red}{0.85\%} & 0.291  & 0.335  & 0.280  & 0.325  & \textcolor{red}{3.50\%} & \textcolor{red}{3.06\%} \\

\midrule
\multirow{4}{*}{\rotatebox[origin=c]{90}{ETTh1}}
& \multicolumn{1}{|c|}{96} & 0.386 & 0.394 & 0.385 & 0.394 & \textcolor{red}{0.09\%} & \textcolor{red}{0.05\%} & 0.399  & 0.419  & 0.401  & 0.420 & \textcolor[HTML]{5CAA82}{-0.48\%} & \textcolor[HTML]{5CAA82}{-0.16\%} & 0.385  & 0.404  & 0.383  & 0.402  & \textcolor{red}{0.67\%} & \textcolor{red}{0.49\%} \\

& \multicolumn{1}{|c|}{192} & 0.436 & 0.423 & 0.436 & 0.423 & \textcolor{red}{0.01\%} & \textcolor{red}{0.01\%} & 0.453  & 0.449  & 0.451  & 0.450 & \textcolor{red}{0.51\%} & \textcolor[HTML]{5CAA82}{-0.27\%} & 0.443  & 0.437  & 0.431  & 0.429  & \textcolor{red}{2.65\%} & \textcolor{red}{1.83\%} \\

& \multicolumn{1}{|c|}{336} & 0.479 & 0.445 & 0.477 & 0.444 & \textcolor{red}{0.37\%} & \textcolor{red}{0.22\%} & 0.507  & 0.480  & 0.502  & 0.477 & \textcolor{red}{0.88\%} & \textcolor{red}{0.57\%} & 0.487  & 0.458  & 0.473  & 0.449  & \textcolor{red}{2.94\%} & \textcolor{red}{1.98\%} \\

& \multicolumn{1}{|c|}{720} & 0.481 & 0.469 & 0.479 & 0.466 & \textcolor{red}{0.40\%} & \textcolor{red}{0.54\%} & 0.532  & 0.502  & 0.505  & 0.490 & \textcolor{red}{4.99\%} & \textcolor{red}{2.42\%} & 0.517  & 0.497  & 0.470  & 0.470  & \textcolor{red}{9.07\%} & \textcolor{red}{5.61\%} \\

\cmidrule(l{10pt}r{10pt}){2-20}
& \multicolumn{1}{|c|}{Avg} & 0.445 & 0.432 & 0.444 & 0.432 & \textcolor{red}{0.23\%} & \textcolor{red}{0.22\%} & 0.473  & 0.463  & 0.465  & 0.459 & \textcolor{red}{1.66\%} & \textcolor{red}{0.70\%} & 0.458  & 0.449  & 0.439  & 0.437  & \textcolor{red}{4.12\%} & \textcolor{red}{2.61\%} \\

\midrule
\multirow{4}{*}{\rotatebox[origin=c]{90}{ETTh2}}
& \multicolumn{1}{|c|}{96} & 0.292 & 0.340 & 0.290 & 0.339 & \textcolor{red}{0.35\%} & \textcolor{red}{0.40\%} & 0.322  & 0.363  & 0.324  & 0.367 & \textcolor[HTML]{5CAA82}{-0.57\%} & \textcolor[HTML]{5CAA82}{-0.91\%}  & 0.299  & 0.350  & 0.295  & 0.346  & \textcolor{red}{1.44\%} & \textcolor{red}{1.04\%} \\

& \multicolumn{1}{|c|}{192} & 0.377 & 0.395 & 0.376 & 0.392 & \textcolor{red}{0.39\%} & \textcolor{red}{0.64\%} & 0.404  & 0.413  & 0.399  & 0.411 & \textcolor{red}{1.18\%} & \textcolor{red}{0.53\%} & 0.380  & 0.399  & 0.377  & 0.398  & \textcolor{red}{0.77\%} & \textcolor{red}{0.17\%} \\

& \multicolumn{1}{|c|}{336} & 0.424 & 0.434 & 0.419 & 0.431 & \textcolor{red}{1.05\%} & \textcolor{red}{0.72\%} & 0.448  & 0.449  & 0.438  & 0.443 & \textcolor{red}{2.26\%} & \textcolor{red}{1.42\%} & 0.423  & 0.432  & 0.419  & 0.431  & \textcolor{red}{0.93\%} & \textcolor{red}{0.37\%} \\

& \multicolumn{1}{|c|}{720} & 0.433 & 0.448 & 0.422 & 0.443 & \textcolor{red}{2.56\%} & \textcolor{red}{1.14\%} & 0.453  & 0.460  & 0.434  & 0.450 & \textcolor{red}{4.27\%} & \textcolor{red}{2.22\%} & 0.429  & 0.446  & 0.419  & 0.441  & \textcolor{red}{2.22\%} & \textcolor{red}{1.18\%} \\

\cmidrule(l{10pt}r{10pt}){2-20}
& \multicolumn{1}{|c|}{Avg} & 0.381 & 0.404 & 0.377 & 0.401 & \textcolor{red}{1.18\%} & \textcolor{red}{0.75\%} & 0.407  & 0.421  & 0.399  & 0.417 & \textcolor{red}{1.99\%} & \textcolor{red}{0.92\%} & 0.383  & 0.407  & 0.378  & 0.404  & \textcolor{red}{1.35\%} & \textcolor{red}{0.69\%} \\

\midrule
\multirow{4}{*}{\rotatebox[origin=c]{90}{Electricity}}
& \multicolumn{1}{|c|}{96} & 0.173 & 0.260 & 0.142 & 0.238 & \textcolor{red}{17.50\%} & \textcolor{red}{8.62\%} & 0.167  & 0.271  & 0.166  & 0.269 & \textcolor{red}{0.80\%} & \textcolor{red}{0.56\%} & 0.148  & 0.240  & 0.134  & 0.232  & \textcolor{red}{9.49\%} & \textcolor{red}{3.56\%} \\

& \multicolumn{1}{|c|}{192} & 0.181 & 0.267 & 0.160 & 0.253 & \textcolor{red}{11.74\%} & \textcolor{red}{5.26\%} & 0.186  & 0.288  & 0.184  & 0.286 & \textcolor{red}{1.15\%} & \textcolor{red}{0.69\%} & 0.165  & 0.256  & 0.152  & 0.248  & \textcolor{red}{7.68\%} & \textcolor{red}{3.19\%} \\

& \multicolumn{1}{|c|}{336} & 0.196 & 0.283 & 0.175 & 0.271 & \textcolor{red}{11.02\%} & \textcolor{red}{4.21\%} & 0.203  & 0.303  & 0.194  & 0.295 & \textcolor{red}{4.58\%} & \textcolor{red}{2.63\%} & 0.178  & 0.270  & 0.167  & 0.267  & \textcolor{red}{6.34\%} & \textcolor{red}{1.45\%} \\

& \multicolumn{1}{|c|}{720} & 0.238 & 0.316 & 0.206 & 0.298 & \textcolor{red}{13.43\%} & \textcolor{red}{5.70\%} & 0.225  & 0.320  & 0.228  & 0.322 & \textcolor[HTML]{5CAA82}{-1.57\%} & \textcolor[HTML]{5CAA82}{-0.61\%} & 0.221  & 0.308  & 0.195  & 0.292  & \textcolor{red}{11.67\%} & \textcolor{red}{5.04\%} \\

\cmidrule(l{10pt}r{10pt}){2-20}
& \multicolumn{1}{|c|}{Avg} & 0.197 & 0.282 & 0.171 & 0.265 & \textcolor{red}{13.33\%} & \textcolor{red}{5.90\%} & 0.195  & 0.295  & 0.193  & 0.293 & \textcolor{red}{1.19\%} & \textcolor{red}{0.80\%} & 0.178  & 0.269  & 0.162  & 0.259  & \textcolor{red}{8.96\%} & \textcolor{red}{3.36\%} \\

\midrule
\multirow{4}{*}{\rotatebox[origin=c]{90}{Weather}}
& \multicolumn{1}{|c|}{96} & 0.155 & 0.203 & 0.151 & 0.197 & \textcolor{red}{2.49\%} & \textcolor{red}{3.34\%} & 0.172  & 0.222  & 0.166  & 0.216 & \textcolor{red}{3.50\%} & \textcolor{red}{2.88\%}  & 0.174  & 0.213  & 0.160  & 0.205  & \textcolor{red}{8.18\%} & \textcolor{red}{4.09\%} \\

& \multicolumn{1}{|c|}{192} & 0.202 & 0.247 & 0.200 & 0.244 & \textcolor{red}{1.33\%} & \textcolor{red}{1.37\%} & 0.229  & 0.268  & 0.216  & 0.259 & \textcolor{red}{5.48\%} & \textcolor{red}{3.66\%} & 0.224  & 0.257  & 0.214  & 0.252  & \textcolor{red}{4.66\%} & \textcolor{red}{1.99\%} \\

& \multicolumn{1}{|c|}{336} & 0.263 & 0.293 & 0.260 & 0.289 & \textcolor{red}{1.23\%} & \textcolor{red}{1.13\%} & 0.284  & 0.305  & 0.274  & 0.299 & \textcolor{red}{3.65\%} & \textcolor{red}{1.87\%} & 0.281  & 0.299  & 0.271  & 0.294  & \textcolor{red}{3.69\%} & \textcolor{red}{1.69\%} \\

& \multicolumn{1}{|c|}{720} & 0.341 & 0.343 & 0.338 & 0.338 & \textcolor{red}{0.99\%} & \textcolor{red}{1.34\%} & 0.357  & 0.354  & 0.351  & 0.349 & \textcolor{red}{1.67\%} & \textcolor{red}{1.29\%} & 0.359  & 0.351  & 0.349  & 0.346  & \textcolor{red}{2.77\%} & \textcolor{red}{1.47\%} \\

\cmidrule(l{10pt}r{10pt}){2-20}
& \multicolumn{1}{|c|}{Avg} & 0.240 & 0.271 & 0.237 & 0.267 & \textcolor{red}{1.37\%} & \textcolor{red}{1.67\%} & 0.261  & 0.287  & 0.252  & 0.281 & \textcolor{red}{3.35\%} & \textcolor{red}{2.30\%} & 0.260  & 0.280  & 0.249  & 0.274  & \textcolor{red}{4.33\%} & \textcolor{red}{2.15\%} \\

\midrule
\multirow{4}{*}{\rotatebox[origin=c]{90}{Traffic}}
& \multicolumn{1}{|c|}{96} & 0.523 & 0.350 & 0.465 & 0.304 & \textcolor{red}{11.07\%} & \textcolor{red}{13.15\%} & 0.590  & 0.315  & 0.588  & 0.314 & \textcolor{red}{0.21\%} & \textcolor{red}{0.25\%} & 0.393  & 0.269  & 0.392  & 0.267  & \textcolor{red}{0.43\%} & \textcolor{red}{0.40\%} \\

& \multicolumn{1}{|c|}{192} & 0.543 & 0.366 & 0.475 & 0.312 & \textcolor{red}{12.50\%} & \textcolor{red}{14.75\%} & 0.617  & 0.329  & 0.612  & 0.326 & \textcolor{red}{0.78\%} & \textcolor{red}{0.85\%} & 0.413  & 0.277  & 0.414  & 0.277  & \textcolor[HTML]{5CAA82}{-0.34\%} & \textcolor{red}{0.13\%} \\

& \multicolumn{1}{|c|}{336} & 0.544 & 0.366 & 0.494 & 0.323 & \textcolor{red}{9.17\%} & \textcolor{red}{11.75\%} & 0.636  & 0.337  & 0.630  & 0.334 & \textcolor{red}{0.98\%} & \textcolor{red}{0.73\%} & 0.424  & 0.283  & 0.427  & 0.283  & \textcolor[HTML]{5CAA82}{-0.66\%} & \textcolor{red}{0.03\%} \\

& \multicolumn{1}{|c|}{720} & 0.590 & 0.395 & 0.532 & 0.344 & \textcolor{red}{9.85\%} & \textcolor{red}{12.83\%} & 0.668  & 0.353  & 0.665  & 0.349 & \textcolor{red}{0.43\%} & \textcolor{red}{1.06\%} & 0.459  & 0.301  & 0.455  & 0.297  & \textcolor{red}{1.03\%} & \textcolor{red}{1.30\%} \\

\cmidrule(l{10pt}r{10pt}){2-20}
& \multicolumn{1}{|c|}{Avg} & 0.550 & 0.369 & 0.492 & 0.321 & \textcolor{red}{10.63\%} & \textcolor{red}{13.12\%} & 0.628  & 0.333  & 0.624  & 0.331 & \textcolor{red}{0.60\%} & \textcolor{red}{0.73\%} & 0.422  & 0.282  & 0.422  & 0.281  & \textcolor{red}{0.13\%} & \textcolor{red}{0.48\%} \\

\midrule
\multicolumn{2}{c|}{Dataset Avg} & \multicolumn{4}{c|}{} & \textcolor{red}{4.13\%} & \textcolor{red}{3.30\%} & \multicolumn{4}{c|}{} & \textcolor{red}{1.62\%} & \textcolor{red}{1.02\%} & \multicolumn{4}{c|}{} & \textcolor{red}{3.81\%} & \textcolor{red}{2.03\%} \\

\bottomrule[1pt]
\end{tabular}
}
\end{table*}

As shown in Table~\ref{tab:generalization_full_ci} and ~\ref{tab:generalization_full_cd}, we can see that TimeSter generally enhances performance across different architectures, datasets, and domains.
Especially, for datasets with strong periodicity and seasonality, such as Electricity, Weather, and Traffic, TimeSter can bring more significant improvements, which is consistent with the characteristics of these datasets.
For these datasets, whether periodicity or seasonality, they are generally based on days, weeks, months, or seasons, which can be captured by time-related features.

However, the performance improvement of iTransformer (Table~\ref{tab:generalization_full_cd}) on the Traffic dataset is minor after combining TimeSter with it.
This may be caused by iTransformer's channel-dependent strategy~\cite{han2023capacity}, where the correlations between different time series variables are modeled.
In contrast, another strategy is the channel-independent strategy, in which models do not model the relationship between different variables, but only treat different variables as independent training samples.
While the cross-variate dependencies between historical observations are modeled by iTransformer, those between historical time-related features are not modeled by TimeSter, which might result in discrepancies between predictions of historical observations and time-related features.
This can also explain the difference in improvement rates between different models: channel-independent models, shown in Table~\ref{tab:generalization_full_ci} (RLinear, FITS, and PatchTST, whose improvement rates are 10.47\%, 10.02\%, and 4.02\% in MSE, respectively), have higher improvement rates while channel-dependent models, shown in Table~\ref{tab:generalization_full_cd} (ModernTCN, TimesNet, and iTransformer, whose improvement rates are 4.13\%, 1.62\%, and 3.81\% in MSE, respectively), have lower improvement rates.
Although differences in improvement rates among models exist, the results still show the strong applicability of TimeSter.
Nonetheless, considering the relationship between different variables in time-related feature modeling is still an explorable topic, and we regard it as a future research direction.

\subsection{Ablation Studies}
\label{sec:ablationstudies}
In ablation studies, we first evaluate the selection of time-related features.
Then, we verify the rationality of the design of the TimeSter module, including the encoder and decoder.
\subsubsection{Ablation of Time-related Features}
The selection of time-related features influences the correct time series pattern recognition, which is particularly important for data with different periodicity and seasonality.
In TimeSter, we generally consider four primary time-related features:
(i) Hour-of-day (H, $0, ..., 23$);
(ii) Day-of-week (D, $0, ..., 6$);
(iii) Month-of-year (M, $0, ..., 11$);
and (iv) Season-of-year (S, $0, ..., 3$).

\begin{table}[htpb]
\caption{Ablation of time-related features. \textit{H} denotes Hour-of-day. \textit{D} indicates Day-of-week. \textit{M} is Month-of-year. \textit{S} indicates Season-of-year. $\times$ means forecasting without TimeSter (i.e., only RLinear). The historical window $L$ is $96$ and the future horizon $T$ is $720$. The best is marked in \textcolor{red}{red}.}
\centering
\label{tab:ablation_timefeatures_main}
\resizebox{1.0\linewidth}{!}{
\begin{tabular}{c|ccccccc}

\toprule[1pt]
\multirow{2}{*}{Features} & \multicolumn{1}{c}{ETTm1} & \multicolumn{1}{c}{ETTm2} & \multicolumn{1}{c}{ETTh1} & \multicolumn{1}{c}{ETTh2} & \multicolumn{1}{c}{Electricity} & \multicolumn{1}{c}{Weather} & \multicolumn{1}{c}{Traffic} \\
\cmidrule(l{10pt}r{10pt}){2-2}\cmidrule(l{10pt}r{10pt}){3-3}\cmidrule(l{10pt}r{10pt}){4-4}\cmidrule(l{10pt}r{10pt}){5-5}\cmidrule(l{10pt}r{10pt}){6-6}\cmidrule(l{10pt}r{10pt}){7-7}\cmidrule(l{10pt}r{10pt}){8-8}
& MSE & MSE & MSE & MSE & MSE & MSE & MSE \\
\toprule[1pt]

$\times$ & 0.486  & 0.407 & 0.480 & 0.422 & 0.253 & 0.364 & 0.643 \\
\midrule

H & \textcolor{red}{0.457}  & \textcolor{red}{0.395} & \textcolor{red}{0.464} & 0.447 & 0.249 & 0.348 & 0.646 \\

H\_D & 0.462  & 0.410 & 0.477 & 0.449 & 0.210 & 0.349 & \textcolor{red}{0.512} \\

H\_M & 0.534  & 0.398 & 0.521 & 0.398 & 0.241 & 0.351 & 0.656 \\

H\_S & 0.479  & 0.398 & 0.508 & 0.417 & 0.236 & \textcolor{red}{0.347} & 0.656 \\

H\_D\_M & 0.547  & 0.412 & 0.515 & 0.382 & 0.202 & 0.355 & 0.535 \\

H\_D\_S  & 0.473  & 0.404 & 0.502 & 0.418 & \textcolor{red}{0.198} & 0.350 & 0.553 \\

H\_M\_S  & 0.559  & 0.399 & 0.519 & \textcolor{red}{0.377} & 0.241 & 0.351 & 0.665 \\

H\_D\_M\_S  & 0.544  & 0.402 & 0.581 & 0.397 & 0.204 & 0.353 & 0.536 \\

\bottomrule[1pt]
\end{tabular}
}
\end{table}
\begin{table*}[tpb]
\caption{Ablation of TimeSter encoder. The historical window $L$ is fixed at $96$ and the average performance of four prediction lengths is reported. The best performers for each dataset are highlighted in \textcolor{red}{red}.}
\centering
\label{tab:ablation_module_design}
\resizebox{1\linewidth}{!}{
\begin{tabular}{c|cc|cc|cc|cc|cc|cc|cc}

\toprule[1pt]
\multicolumn{1}{c}{Dataset} &
\multicolumn{2}{c}{ETTm1}  & \multicolumn{2}{c}{ETTm2} & \multicolumn{2}{c}{ETTh1} & \multicolumn{2}{c}{ETTh2} & \multicolumn{2}{c}{Electricity} & \multicolumn{2}{c}{Weather} & \multicolumn{2}{c}{Traffic}  \\
\cmidrule(l{10pt}r{10pt}){1-1}\cmidrule(l{10pt}r{10pt}){2-3}\cmidrule(l{10pt}r{10pt}){4-5}\cmidrule(l{10pt}r{10pt}){6-7}\cmidrule(l{10pt}r{10pt}){8-9}\cmidrule(l{10pt}r{10pt}){10-11}\cmidrule(l{10pt}r{10pt}){12-13}\cmidrule(l{10pt}r{10pt}){14-15}
\multicolumn{1}{c}{Metric} & MSE & MAE & MSE & MAE & MSE & MAE & MSE & MAE & MSE & MAE & MSE & MAE & MSE & MAE \\
\toprule[1pt]

TimeLinear & \textcolor{red}{0.385} & \textcolor{red}{0.395} & \textcolor{red}{0.273} & \textcolor{red}{0.315} & \textcolor{red}{0.432} & \textcolor{red}{0.426} & \textcolor{red}{0.358} & \textcolor{red}{0.389} & \textcolor{red}{0.165} & \textcolor{red}{0.259} & \textcolor{red}{0.251} & \textcolor{red}{0.276} & 0.480 & \textcolor{red}{0.304} \\

w / o ReLU & 0.386 & \textcolor{red}{0.395} & \textcolor{red}{0.273} & \textcolor{red}{0.315} & 0.433 & \textcolor{red}{0.426} & 0.369 & 0.395 & 0.172 & 0.265 & \textcolor{red}{0.251} & \textcolor{red}{0.276} & \textcolor{red}{0.477} & 0.312 \\

w / o Conv1d & 0.387 & 0.396 & 0.275 & 0.318 & 0.435 & 0.427 & 0.367 & 0.394 & 0.170 & 0.263 & 0.252 & 0.278 & 0.484 & 0.310 \\

w / o LayerNorm & 0.386 & 0.396 & 0.274 & 0.317 & 0.434 & \textcolor{red}{0.426} & 0.369 & 0.395 & 0.168 & 0.262 & \textcolor{red}{0.251} & \textcolor{red}{0.276} & 0.480 & 0.306 \\

One Hidden Layer & 0.386 & \textcolor{red}{0.395} & \textcolor{red}{0.273} & 0.316 & 0.434 & 0.427 & 0.365 & 0.394 & 0.167 & 0.261 & \textcolor{red}{0.251} & \textcolor{red}{0.276} & 0.482 & 0.308 \\

Zero Hidden Layer & 0.388 & 0.396 & 0.278 & 0.322 & 0.440 & 0.429 & 0.370 & 0.397 & 0.184 & 0.274 & 0.253 & 0.278 & 0.525 & 0.350 \\

\bottomrule[1pt]
\end{tabular}
}
\end{table*}
\begin{table*}[tpb]
\caption{Ablation of TimeSter decoder. Both $f$ and $g$ are linear projectors.
$\mathbf{X}\in\mathbb{R}^{L\times V}$ denotes historical observations of $V$ variables. $\mathbf{U}\in\mathbb{R}^{L\times r}$ denotes the time-related features corresponding to the historical data, and $\mathbf{P}\in\mathbb{R}^{T\times r}$ denotes the time-related features corresponding to the future data.
$q_\theta(\cdot)$ is the TimeSter encoder.
The best performers for each dataset are highlighted in \textcolor{red}{red}.}
\centering
\label{tab:ablation_embedding_mode}
\resizebox{1.0\linewidth}{!}{
\begin{tabular}{c|c|cc|cc|cc|cc|cc|cc|cc}

\toprule[1pt]
\multirow{2}{*}{Variant} & \multirow{2}{*}{Mode} &
\multicolumn{2}{c}{ETTm1}  & \multicolumn{2}{c}{ETTm2} & \multicolumn{2}{c}{ETTh1} & \multicolumn{2}{c}{ETTh2} & \multicolumn{2}{c}{Electricity} & \multicolumn{2}{c}{Weather} & \multicolumn{2}{c}{Traffic}  \\
\cmidrule(l{10pt}r{10pt}){3-4}\cmidrule(l{10pt}r{10pt}){5-6}\cmidrule(l{10pt}r{10pt}){7-8}\cmidrule(l{10pt}r{10pt}){9-10}\cmidrule(l{10pt}r{10pt}){11-12}\cmidrule(l{10pt}r{10pt}){13-14}\cmidrule(l{10pt}r{10pt}){15-16}
& & MSE & MAE & MSE & MAE & MSE & MAE & MSE & MAE & MSE & MAE & MSE & MAE & MSE & MAE \\
\toprule[1pt]

1 & $f(\mathbf{X})$ & 0.412 & 0.406 & 0.286 & 0.327 & 0.446 & 0.433 & 0.377 & 0.399 & 0.215 & 0.291 & 0.273 & 0.291 & 0.623 & 0.371 \\

2 &$q_\theta(\mathbf{P})$ & 0.419 & 0.426 & 0.292 & 0.328 & 0.451 & 0.453 & 0.421 & 0.429 & 0.196 & 0.295 & 0.265 & 0.289 & 0.653 & 0.356 \\

3 &$f(q_\theta(\mathbf{U}))$ & 0.419 & 0.426 & 0.293 & 0.328 & 0.451 & 0.454 & 0.390 & 0.411 & 0.195 & 0.294 & 0.266 & 0.290 & 0.657 & 0.356 \\

4 &$f(\mathbf{X})+q_\theta(\mathbf{P})$ & 0.386 & 0.396 & \textcolor{red}{0.272} & \textcolor{red}{0.315} & 0.436 & 0.428 & 0.379 & 0.402 & 0.167 & 0.261 & \textcolor{red}{0.251} & \textcolor{red}{0.276} & \textcolor{red}{0.479} & \textcolor{red}{0.303} \\

5 &$f(\mathbf{X}+q_\theta(\mathbf{U}))$ & 0.386 & \textcolor{red}{0.395} & 0.275 & 0.318 & 0.434 & 0.427 & 0.367 & 0.394 & 0.168 & 0.264 & \textcolor{red}{0.251} & \textcolor{red}{0.276} & 0.481 & 0.307 \\

6 &$f(\mathbf{X})+g(q_\theta(\mathbf{U}))$ & \textcolor{red}{0.385} & \textcolor{red}{0.395} & 0.273 & \textcolor{red}{0.315} & \textcolor{red}{0.432} & \textcolor{red}{0.426} & \textcolor{red}{0.358} & \textcolor{red}{0.389} & \textcolor{red}{0.165} & \textcolor{red}{0.259} & \textcolor{red}{0.251} & \textcolor{red}{0.276} & 0.480 & 0.304 \\

\bottomrule[1pt]
\end{tabular}
}
\end{table*}

Results, which are consistent with the analysis of dataset characteristics~\cite{lai2018modeling,zhou2021informer,lin2024cyclenet}, are illustrated in Table~\ref{tab:ablation_timefeatures_main}.
In general, ETTm1, ETTm2, and ETTh1 have strong daily cycles, while Traffic shows significant daily and weekly cycles.
ETTh2, Electricity, and Weather exhibit more significant seasonality than other datasets.
Notably, the Weather dataset, which is supposed to have strong seasonality, does not achieve significant gains in seasonal modeling (features M and S).
This is because the time span of the Weather dataset is only one year, and the data used as the training set is only eight months, as shown in Table~\ref{tab:dataset}.
As a result, the model can not fully capture all seasonal characteristics.
However, TimeSter still achieves performance improvement under this limitation, which proves that TimeSter has strong generalization ability.
The above results also illustrate the importance of time-related feature selection, which is ignored by previous methods~\cite{wang2024rethinking,liu2024autotimes,liu2024itransformer}.

\subsubsection{Ablation of TimeSter Encoder}
To further validate the design of TimeSter, we conduct extensive experiments on the effectiveness of each module in the TimeSter encoder.
Results are shown in Table~\ref{tab:ablation_module_design}:
(i) Removing ReLU significantly impacts datasets with more complex time-related feature dependencies, such as Electricity and ETTh2, where seasonality is considered.
This highlights the importance of nonlinearity in learning intricate mappings.
(ii) The Conv1d layer captures local feature correlations via convolution and global temporal dependencies via channel mixing.
This not only enriches the semantic information of each feature but also enhances the temporal dependence of encoding.
Therefore, compared to other modules, its removal will result in significant performance degradation.
(iii) LayerNorm’s stabilizing effect could provide a subtle advantage by ensuring consistent scaling, which aids the model’s learning dynamics across varying time series patterns.
(iv) Varying the hidden layer depth reveals that a deeper architecture achieves better performance, especially when compared to the \textit{Zero Hidden Layer} model, which shows a substantial performance drop.
This emphasizes the importance of model depth for capturing complex temporal patterns.
The \textit{One Hidden Layer} variant, while the performance is closer to the TimeSter, is still slightly inferior, indicating that additional depth enhances the model's representational power.
However, the benefits of increasing the number of layers gradually decrease.
Hence, for efficiency and accuracy considerations, the final model only considers two hidden layers.
Overall, the full architecture of the TimeSter encoder is rational, as each component contributes to performance improvement.

\subsubsection{Ablation of TimeSter Decoder}
Our TimeSter module leverages time-related features by projecting the time-related observations generated by the encoder to future observations and adding them to the results generated by the backbone model.
However, there are other embedding modes, e.g., adding the historical time-related observations with historical observations or adding the future time-related observations with the results generated by the backbone model.
To demonstrate the effectiveness of our mode, we make a comprehensive comparison with all these variants in this section, including:
\begin{itemize}
    \item Variant 1: $f(\mathbf{X})$ (i.e., RLinear), where we predict with historical observations using a single linear layer and simplified RevIN;
    \item Variant 2: $q_\theta(\mathbf{P})$, where we predict with the future time-related observations generated by TimeSter encoder;
    \item Variant 3: $f(q_\theta(\mathbf{U}))$, where we predict with a linear projector that takes the historical time-related observations as the input;
    \item Variant 4: $f(\mathbf{X})+q_\theta(\mathbf{P})$, where we add the output of the linear projector and the future time-related observations;
    \item Variant 5: $f(\mathbf{X}+q_\theta(\mathbf{U}))$, where we add the input historical observations and time-related observations;
    \item Variant 6: $f(\mathbf{X})+g(q_\theta(\mathbf{U}))$, our TimeLinear, where we predict with two linear layers ($f$ and $g$) that take historical observations and historical time-related observations as input respectively and add the results up.
\end{itemize}

The results are shown in Table~\ref{tab:ablation_embedding_mode}, where we can draw some interesting conclusions:
(i) Predicting with time-related features surpasses predicting with historical observations in some datasets (Electricity and Weather), deriving from the comparison of Variant 1, 2, and 3.
(ii) Historical time-related features and future time-related features exhibit similar effectiveness, deriving from Variant 2, 3 and Variant 4, 5.
(iii) TimeLinear achieves the best overall performance.
These findings not only prove the effectiveness of our time-related feature modeling but also verify the rationality of our dynamic projection strategy.
Moreover, our experiments in the next section will show that as the historical window increases, our dynamic projection strategy is particularly more advantageous than others.

\subsection{More Analysis}
\label{sec:moreanalysis}
In this section, we begin with further experiments from the last section, evaluating the effectiveness of TimeSter under longer historical windows.
Then, we explore the mechanism of time-related features visually.
Furthermore, we use the quantitative indicator ACF to explore the relationship between time-related features and the characteristics of the corresponding time series.
Finally, we conduct the robustness and hyperparameter analysis.
\subsubsection{Longer Historical Windows}
\begin{figure}[htpb]
  \centering
  \includegraphics[width=1\columnwidth]{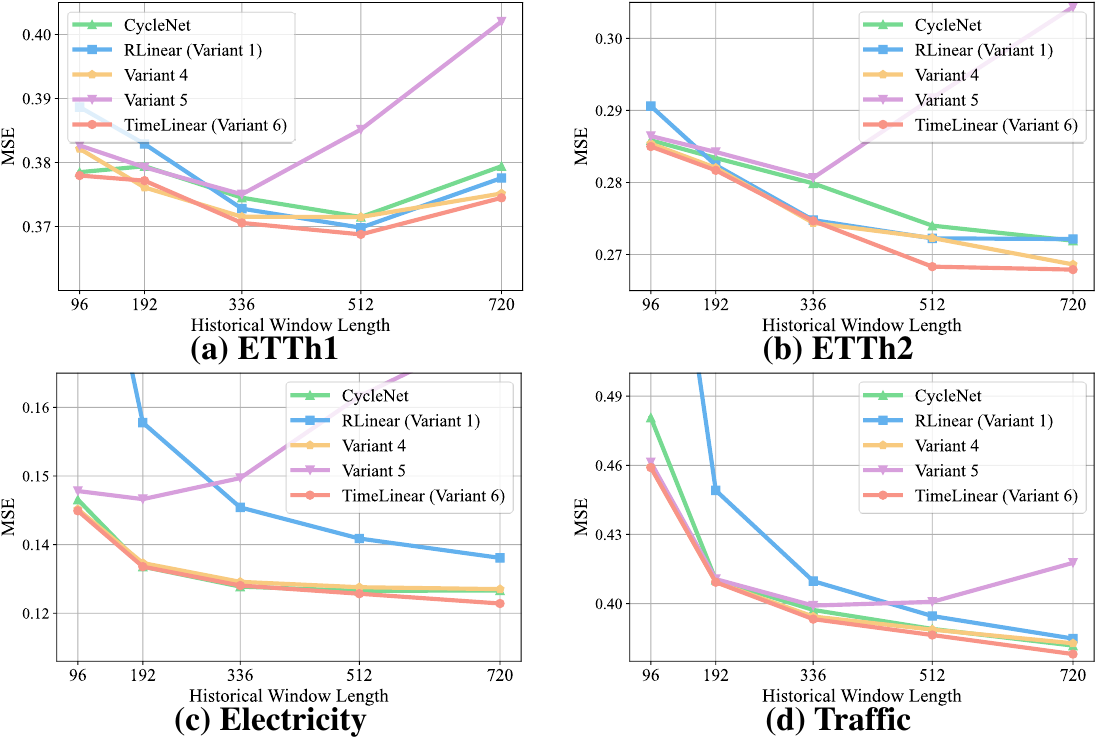}
  \caption{
  Performance promotion with longer historical windows.
  The forecasting length is $96$.}
  \label{fig:longer}
\end{figure}

In long-term time series forecasting, the historical window length is an important hyperparameter, which determines the richness of temporal information within input data.
Both theory~\cite{box1968some} and practice~\cite{Yuqietal-2023-PatchTST,liu2024itransformer,lin2024cyclenet} have proven that a longer historical window often leads to better forecasting performance.
For time-related features, a longer historical window indicates richer time-related information, which is supposed to be useful for more accurate forecasting.
Figure~\ref{fig:longer} illustrates the performance of CycleNet~\cite{lin2024cyclenet} and our variants in Table~\ref{tab:ablation_embedding_mode} under different historical window lengths, where some interesting findings could be drawn:
(i) TimeLinear outperforms baselines and other variants at almost all historical window lengths.
(ii) Variant 5 struggles to provide better prediction over a longer historical window.
This is because longer historical windows also mean greater noise, and simply adding time-related observations with historical observations will lead to mutual interference of noise~\cite{pereira2023comparative,wang2024rethinking}, which will eventually lead to worse prediction results.
(iii) While Variant 4 is close to TimeLinear under shorter historical windows, it gradually loses ground as the historical window increases, for it only relies on future time stamps and cannot utilize richer input time stamp information.
(iv) The growth of the historical window also leads to a weakening of the gain brought by the time stamp, as richer historical observation information contains more cycle and seasonal information, which dilutes the benefits of time-related information modeling.
Such a phenomenon is worthy of further exploration.


\subsubsection{Time-related Features Visualization}
\begin{figure}[htpb]
  \centering
  \includegraphics[width=0.99\columnwidth]{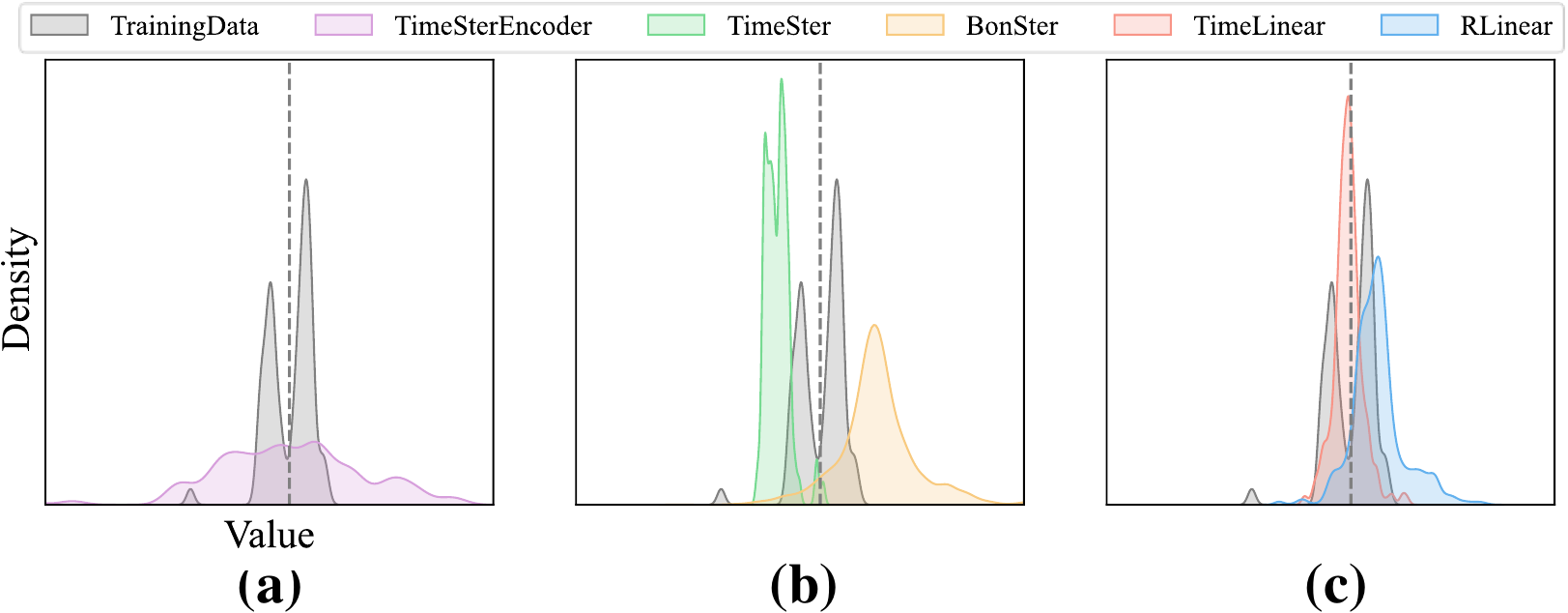}
  \caption{
  Data distribution of the Traffic dataset at noon on Mondays.
  The variable index is $26$. Equipping TimeSter enables the model to make predictions that are closer to the true distribution.
  }
  \label{fig:visual}
\end{figure}

To reveal the relationship between time stamps and variable observations, we visualize the learned time-related embedding and the output of different modules, shown in Figure~\ref{fig:visual}, where the distribution is depicted.
On the left figure, we find that the embedding of the TimeSter encoder tends to be more uniform, indicating that it pays more attention to learning diverse encoding.
The next figure shows that TimeSter learns the shape of the data distribution well and corrects BonSter's high prediction values.
The right figure also proves this point.
After combining TimeSter, the model can make predictions that are more in line with the characteristics of that time point, that is, they are distributed along the mean (shown by the dotted line).

\subsubsection{Correlations between Datasets and Time-related Features}
Different datasets have different periodicities and seasonalities, which in turn affect the selection of time-related features.
For example, hour-of-day (H) often corresponds to daily periodicity, while season-of-year (S) corresponds to seasonality.
To analyze the periodicity and seasonality of different datasets, and explain the relationship between these properties and time-related feature selection, we introduce the Autocorrelation Function (ACF)~\cite{madsen2007time,lin2024cyclenet}.

The Autocorrelation Function (ACF) is a statistical tool used to measure the correlation of a time series with its lagged versions.
It provides insight into the degree of similarity between values separated by lags, which helps identify patterns, seasonality, or periodicity in time series.
The autocorrelation at lag \( k \) is defined as the correlation between observations separated by \( k \) time steps.
For a univariate time series \( X \) with mean \( \mu \), the ACF \( \rho_k \) at lag \( k \) is computed as:

\[
\rho_k = \frac{\sum_{t=1}^{N-k} (x_t - \mu)(x_{t+k} - \mu)}{\sum_{t=1}^{N} (x_t - \mu)^2}\,,
\]
where \( N \) is the number of observations, \( \mu \) indicates the mean of the time series \( X \), \( x_t \) denotes the value of the time series at time step \( t \), and \( x_{t+k} \) is the value at time step \( t + k \), representing a lag of \( k \).
The values of \( \rho_k \) range from \(-1\) to \(1\), where \( 1 \) denotes perfect positive correlation at lag \( k \), \( 0 \) indicates no correlation at lag \( k \), and \( -1 \) means perfect negative correlation at lag \( k \).
\begin{table*}[tpb]
\caption{Performance changes after adding minute-of-hour (Min) to datasets with minute granularity. \textcolor{red}{$\uparrow$} indicates improved performance, \textcolor{blue}{$=$} indicates the same performance, and \textcolor[HTML]{5CAA82}{$\downarrow$} denotes decreasing performance. The best performer under different time-related feature combinations is \textbf{bold}. The historical window $L$ is $96$ and the future horizon $T$ is $720$.}
\centering
\label{tab:ablation_timefeatures_min}
\resizebox{0.99\linewidth}{!}{
\begin{tabular}{c|cccc|cccc|cccccccc}

\toprule[1pt]
\multirow{2}{*}{Time Features} & \multicolumn{2}{c}{ETTm1} & \multicolumn{2}{c|}{+Min} & \multicolumn{2}{c}{ETTm2} & \multicolumn{2}{c|}{+Min} & \multicolumn{2}{c}{Weather} & \multicolumn{2}{c}{+Min} \\
\cmidrule(l{10pt}r{10pt}){2-3}\cmidrule(l{10pt}r{10pt}){4-5}\cmidrule(l{10pt}r{10pt}){6-7}\cmidrule(l{10pt}r{10pt}){8-9}\cmidrule(l{10pt}r{10pt}){10-11}\cmidrule(l{10pt}r{10pt}){12-13}
& MSE & MAE & MSE & MAE & MSE & MAE & MSE & MAE & MSE & MAE & MSE & MAE \\
\toprule[1pt]

H & \textbf{0.457} & \textbf{0.434} & \textbf{0.456}\textcolor{red}{$\uparrow$} & \textbf{0.433}\textcolor{red}{$\uparrow$} & \textbf{0.395} & \textbf{0.390} & \textbf{0.395}\textcolor{blue}{$=$} & \textbf{0.390}\textcolor{blue}{$=$} & 0.348 & 0.344 & 0.348\textcolor{blue}{$=$} & 0.344\textcolor{blue}{$=$} \\

H\_D & 0.462 & 0.437 & 0.463\textcolor[HTML]{5CAA82}{$\downarrow$} & 0.438\textcolor[HTML]{5CAA82}{$\downarrow$} & 0.410 & 0.400 & 0.413\textcolor[HTML]{5CAA82}{$\downarrow$} & 0.402\textcolor[HTML]{5CAA82}{$\downarrow$} & 0.349 & 0.345 & 0.349\textcolor{blue}{$=$} & 0.344\textcolor{red}{$\uparrow$} \\

H\_M & 0.534 & 0.480 & 0.529\textcolor{red}{$\uparrow$} & 0.479\textcolor{red}{$\uparrow$} & 0.398 & 0.393 & 0.401\textcolor[HTML]{5CAA82}{$\downarrow$} & 0.395\textcolor[HTML]{5CAA82}{$\downarrow$} & 0.351 & 0.346 & 0.349\textcolor{red}{$\uparrow$} & 0.345\textcolor{red}{$\uparrow$} \\

H\_S & 0.479 & 0.449 & 0.483\textcolor[HTML]{5CAA82}{$\downarrow$} & 0.452\textcolor[HTML]{5CAA82}{$\downarrow$} & 0.398 & 0.392 & 0.399\textcolor[HTML]{5CAA82}{$\downarrow$} & 0.391\textcolor{red}{$\uparrow$} & \textbf{0.347} & \textbf{0.342} & \textbf{0.347}\textcolor{blue}{$=$} & \textbf{0.343}\textcolor[HTML]{5CAA82}{$\downarrow$} \\

H\_D\_M & 0.547 & 0.488 & 0.525\textcolor{red}{$\uparrow$} & 0.479\textcolor{red}{$\uparrow$} & 0.412 & 0.402 & 0.412\textcolor{blue}{$=$} & 0.402\textcolor{blue}{$=$} & 0.355 & 0.350 & 0.359\textcolor[HTML]{5CAA82}{$\downarrow$} & 0.351\textcolor[HTML]{5CAA82}{$\downarrow$} \\

H\_D\_S & 0.473 & 0.445 & 0.472\textcolor{red}{$\uparrow$} & 0.445\textcolor{blue}{$=$} & 0.404 & 0.394 & 0.405\textcolor[HTML]{5CAA82}{$\downarrow$} & 0.397\textcolor[HTML]{5CAA82}{$\downarrow$} & 0.350 & 0.345 & 0.350\textcolor{blue}{$=$} & 0.345\textcolor{blue}{$=$} \\

H\_M\_S & 0.559 & 0.494 & 0.544\textcolor{red}{$\uparrow$} & 0.487\textcolor{red}{$\uparrow$} & 0.399 & 0.394 & 0.404\textcolor[HTML]{5CAA82}{$\downarrow$} & 0.398\textcolor[HTML]{5CAA82}{$\downarrow$} & 0.351 & 0.347 & 0.351\textcolor{blue}{$=$} & 0.348\textcolor[HTML]{5CAA82}{$\downarrow$} \\

H\_D\_M\_S & 0.544 & 0.487 & 0.546\textcolor[HTML]{5CAA82}{$\downarrow$} & 0.488\textcolor[HTML]{5CAA82}{$\downarrow$} & 0.402 & 0.395 & 0.406\textcolor[HTML]{5CAA82}{$\downarrow$} & 0.397\textcolor[HTML]{5CAA82}{$\downarrow$} & 0.353 & 0.347 & 0.353\textcolor{blue}{$=$} & 0.348\textcolor[HTML]{5CAA82}{$\downarrow$} \\
\midrule
\textcolor{red}{$\uparrow$}:\textcolor{blue}{$=$}:\textcolor[HTML]{5CAA82}{$\downarrow$} & \multicolumn{4}{c|}{\textcolor{red}{9}:\textcolor{blue}{1}:\textcolor[HTML]{5CAA82}{6}} & \multicolumn{4}{c|}{\textcolor{red}{1}:\textcolor{blue}{4}:\textcolor[HTML]{5CAA82}{11}} & \multicolumn{4}{c}{\textcolor{red}{3}:\textcolor{blue}{8}:\textcolor[HTML]{5CAA82}{5}} \\

\bottomrule[1pt]
\end{tabular}
}
\end{table*}
\begin{table*}[htpb]
\caption{Time-related features for different datasets and prediction lengths. Min: Minute-of-hour, H: Hour-of-day, D: Day-of-week, M: Month-of-year, S: Season-of-year.}
\centering
\label{tab:features}
\resizebox{0.99\linewidth}{!}{
\begin{tabularx}{\textwidth}{c|>{\centering\arraybackslash}X|>{\centering\arraybackslash}X|>{\centering\arraybackslash}X|>{\centering\arraybackslash}X|>{\centering\arraybackslash}X|>{\centering\arraybackslash}X|>{\centering\arraybackslash}X}

\toprule[1pt]
\multicolumn{1}{c|}{{Dataset}} & ETTm1  & ETTm2 & ETTh1 & ETTh2 & Electricity & Weather & Traffic \\

\toprule[1pt]
96 & Min\_H & H & H & H & H\_D & H\_S & H\_D \\
192 & Min\_H & H & H & H & H\_D & H\_S & H\_D \\
336 & Min\_H & H & H & H\_D\_M & H\_D\_S & H\_S & H\_D \\
720 & Min\_H & H & H & H\_M\_S & H\_D\_S & H\_S & H\_D \\

\bottomrule[1pt]
\end{tabularx}
}
\end{table*}

\begin{figure}[htpb]
  \centering
  \includegraphics[width=1\columnwidth]{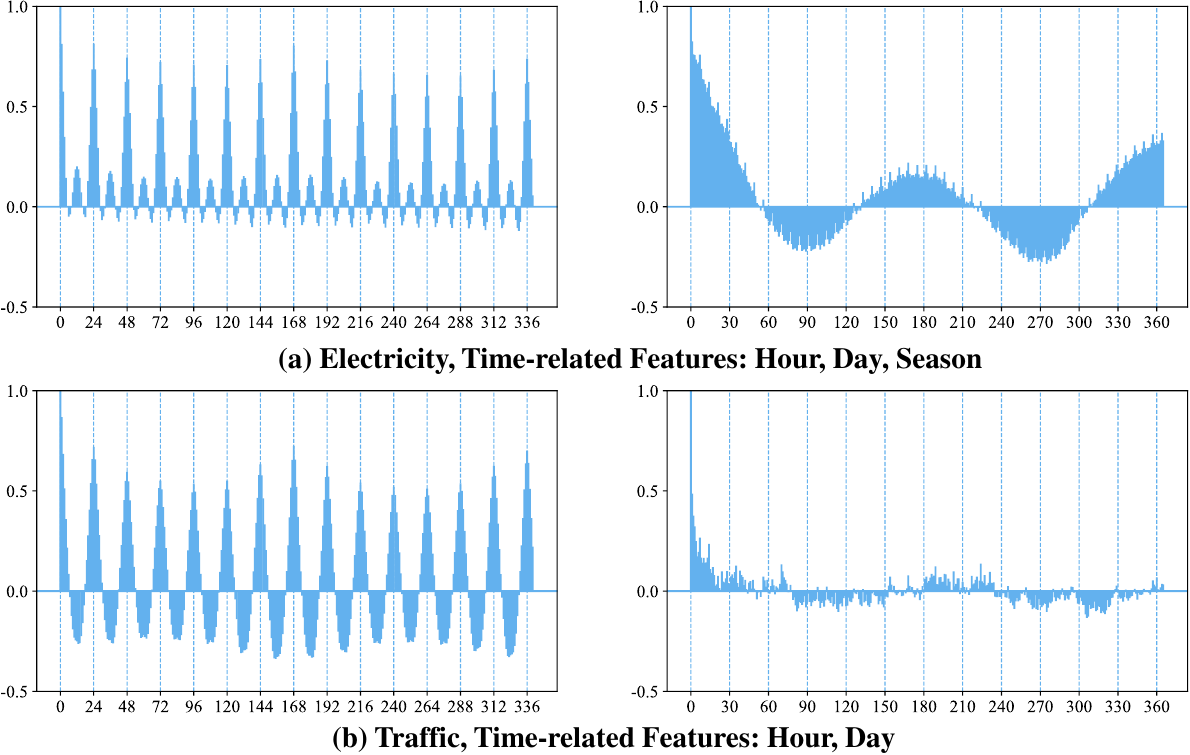}
  \caption{ACF of Electricity and Traffic datasets. The variable index is 28 for both datasets. Granularity is hour for the left figure and day for the right figure. The selection of time-related features is highly consistent with the periodic and seasonal characteristics of the time series itself.}
  \label{fig:acf}
\end{figure}

For an ACF curve, with $k$ as the horizontal axis, its periodic peaks indicate the period of the time series.
For instance, a peak that occurs every 12 time steps means that the original sequence has a component with a period of 12.
If we downsample the original time series, ACF can also be used to explore the seasonality of the original series.
Figure~\ref{fig:acf} illustrates the ACF curve of the Electricity and Traffic dataset respectively, where the granularity is hour for the left figure and day for the right one.
From the left figure, we can see that both datasets show strong daily periodicity, which is reflected in the peaks of the ACF curve that appear every 24 hours.
In addition, both datasets also show a certain weekly periodicity, which is reflected in the larger peaks that appear every 168 hours (7 days).
From the right figure, where the granularity is day, we can find that the Electricity dataset shows a strong partition distribution phenomenon, that is, there is a peak approximately every 90 days (3 months), and there are long-term similar distributions on both sides of the peak.
In contrast, such a phenomenon does not occur in the Traffic dataset.
This indicates that the Electricity dataset has more significant seasonality than the Traffic dataset.
These characteristics correspond one-to-one with the selection of time-related features, which further proves the importance and effectiveness of time-related features for extracting the periodicity and seasonality of time series.
\begin{figure}[htpb]
  \centering
  \includegraphics[width=1\columnwidth]{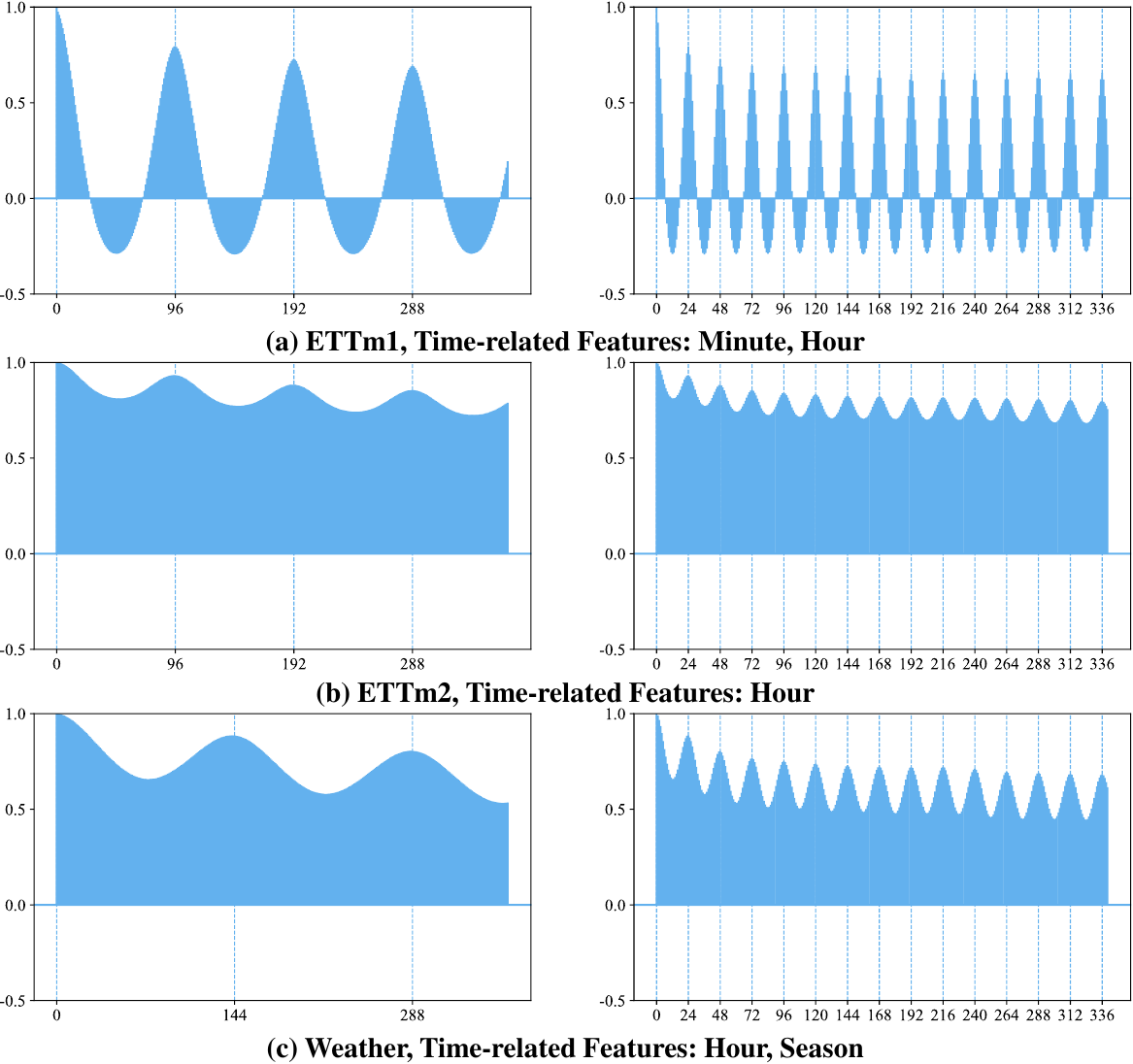}
  \caption{ACF of ETTm1, ETTm2, and Weather. Granularity is minute for the left figure (15 minutes for ETTm1, ETTm2, and 10 minutes for Weather) and hour for the right figure. These datasets show stronger hourly periodicity.}
  \label{fig:min}
\end{figure}

\subsubsection{Minute-of-Hour for Datasets with Minute Granularity}
\label{sec:moh}
In ablation studies, we evaluate the influence of time-related features under the hour granularity.
However, the minute-of-hour (Min) feature for datasets with minute granularity (ETTm1, ETTm2, and Weather) needs further analysis.
In Table~\ref{tab:ablation_timefeatures_min}, we compare the results after adding the minute-of-hour (Min) feature with the original results in Table~\ref{tab:ablation_timefeatures_main}.
The results show that although the minute feature improves the best performance of ETTm1, in most cases, it not only does not bring improvement but deteriorates model performance.
We attribute the reason to the fact that these time series are not strongly correlated with the minute.
To prove this, we illustrate the Autocorrelation Function (ACF) of ETTm1, ETTm2, and Weather under different granularity in Figure~\ref{fig:min}.
We can see that these datasets clearly exhibit a stronger hourly periodicity, for the peak always occurs in a 24-hour cycle.
ETTm1 exhibits a slightly different distribution, which explains why its performance improves after adding the minute feature.
The above findings further demonstrate that our modeling of time stamps well reflects the intrinsic cyclical and seasonal characteristics of time series.

\subsubsection{Time-related Feature Choice for Different Datasets and Prediction Lengths}
Due to the differences in granularity and temporal characteristics of datasets and the prediction time span, the optimal time-related features are different for different datasets and prediction lengths.
In Table~\ref{tab:features}, we illustrate the optimal time-related features of each dataset and prediction length under TimeLinear.

Generally, the choice of time-related features is consistent with the characteristics of the dataset itself, which are shown in both Figure~\ref{fig:acf} and Figure~\ref{fig:min}.
As the prediction length increases, time-related features with longer time spans (e.g., month-of-year and season-of-year) become more effective for time series with stronger seasonality, as shown in the ETTh2 and Electricity with 336 or 720 prediction length.
The same phenomenon also exists for longer historical windows.
It should be noted that the above results are only obtained from experiments under our TimeLinear model.
For other architectures (e.g., PatchTST~\cite{Yuqietal-2023-PatchTST}, ModernTCN~\cite{luo2024moderntcn}), although the overall selection of time-related features is consistent with TimeLinear, due to the different abilities of models in capturing cross-time and cross-variate dependencies in historical observations, the optimal selection of time-related features might also vary slightly.

\subsubsection{Robustness Analysis}
Robustness is another important indicator to measure model performance.
Table 10 shows the mean and standard deviation of TimeLinear under multiple random trainings.
It can be seen that TimeLinear has strong stability as it shows a standard deviation close to 0 in almost all settings.
This further indicates the practicality of our model.
\begin{table*}[htpb]
\caption{Robustness of TimeLinear performance. The average results from three random seeds \{2020, 2021, 2022\} are reported.}
\centering
\label{tab:robustness}
\resizebox{1.0\linewidth}{!}{
\begin{tabular}{c|c|c|c|c|c|c|c|c}

\toprule[1pt]
\multicolumn{2}{c|}{\diagbox{Horizon}{Dataset}} & ETTm1  & ETTm2 & ETTh1 & ETTh2 & Electricity & Weather & Traffic \\

\toprule[1pt]

\multirow{2}{*}{96}
&MSE & 0.325$\pm$0.000 & 0.167$\pm$0.000 & 0.378$\pm$0.001 & 0.285$\pm$0.001 & 0.140$\pm$0.000 & 0.166$\pm$0.000 & 0.459$\pm$0.000 \\
&MAE & 0.364$\pm$0.000 & 0.249$\pm$0.000 & 0.391$\pm$0.000 & 0.335$\pm$0.000 & 0.234$\pm$0.000 & 0.212$\pm$0.000 & 0.293$\pm$0.000 \\

\midrule
\multirow{2}{*}{192}
&MSE & 0.365$\pm$0.000 & 0.233$\pm$0.000 & 0.424$\pm$0.001 & 0.373$\pm$0.001 & 0.155$\pm$0.000 & 0.218$\pm$0.000 & 0.467$\pm$0.000 \\
&MAE & 0.381$\pm$0.001 & 0.291$\pm$0.000 & 0.418$\pm$0.001 & 0.390$\pm$0.001 & 0.247$\pm$0.000 & 0.256$\pm$0.001 & 0.298$\pm$0.000 \\

\midrule
\multirow{2}{*}{336}
&MSE & 0.395$\pm$0.000 & 0.295$\pm$0.000 & 0.463$\pm$0.001 & 0.398$\pm$0.001 & 0.169$\pm$0.000 & 0.272$\pm$0.000 & 0.481$\pm$0.000 \\
&MAE & 0.401$\pm$0.000 & 0.331$\pm$0.000 & 0.438$\pm$0.000 & 0.418$\pm$0.002 & 0.265$\pm$0.000 & 0.294$\pm$0.000 & 0.305$\pm$0.000 \\

\midrule
\multirow{2}{*}{720}
&MSE & 0.456$\pm$0.000 & 0.395$\pm$0.000 & 0.464$\pm$0.001 & 0.377$\pm$0.009 & 0.198$\pm$0.000 & 0.347$\pm$0.000 & 0.512$\pm$0.000 \\
&MAE & 0.433$\pm$0.000 & 0.390$\pm$0.000 & 0.456$\pm$0.000 & 0.412$\pm$0.003 & 0.290$\pm$0.000 & 0.342$\pm$0.000 & 0.320$\pm$0.000 \\

\bottomrule[1pt]
\end{tabular}
}
\end{table*}

\begin{figure}[htpb]
  \centering
  \includegraphics[width=0.96\columnwidth]{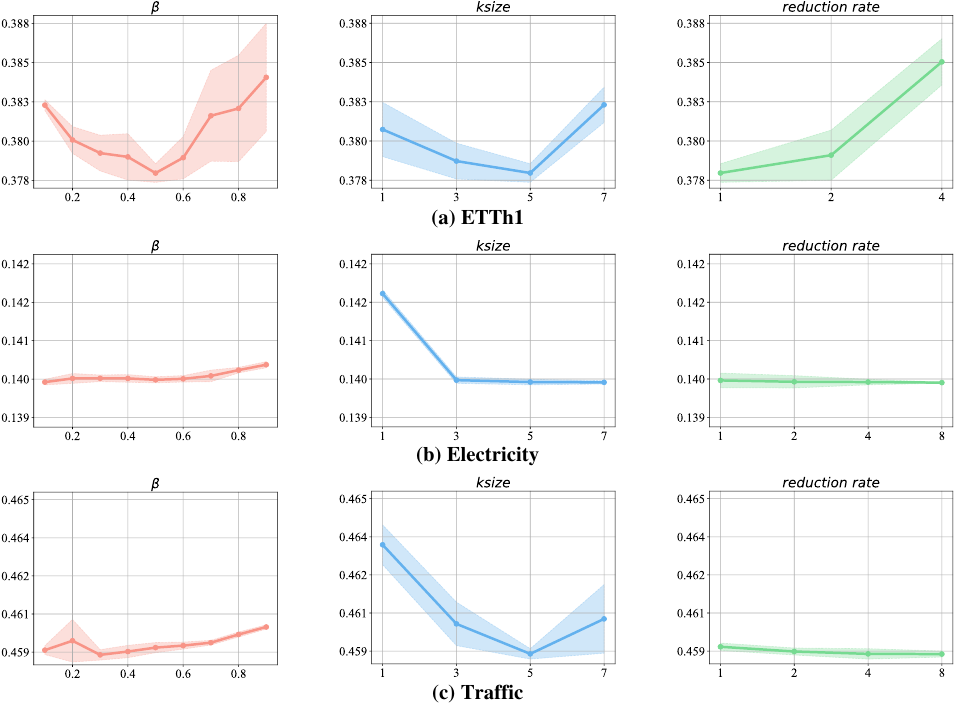}
  \caption{Hyperparameter sensitivity of trade-off coefficient $\mathbf{\beta}$, convolution kernel size \textit{ksize}, and hidden layer size \textit{reduction rate} for ETTh1, Electricity, and Traffic.}
  \label{fig:hypersensi}
\end{figure}

\subsubsection{Hyperparameter Sensitivity}
In addition to the basic learning rate, batch size, etc., various hyperparameters affect the performance of TimeLinear, mainly including the trade-off coefficient $\mathbf{\beta}$, convolution kernel size \textit{ksize}, and hidden layer size.
In this section, we detail the sensitivity of TimeLinear to these hyperparameters on different datasets.
Note that here we only consider the size of the first hidden layer, because the size of the second hidden layer is always set to the number of variables in the dataset.
In addition, for the size of the first hidden layer, we express it as an \textit{reduction rate} relative to the number of variables, e.g., \textit{reduction rate} $=2$ for Traffic with 862 variables means the hidden size is 431.

The results in Figure~\ref{fig:hypersensi} show that $\mathbf{\beta}$ and \textit{ksize} have significant impacts on each dataset, for they directly affect the fusion of features and prediction results, which is crucial for effective time-related feature modeling.
The size of the hidden layer determined by the \textit{reduction rate} is related to the number of variables in the dataset, so its impact varies from dataset to dataset.
\begin{figure}[tpb]
  \centering
  \includegraphics[width=0.99\columnwidth]{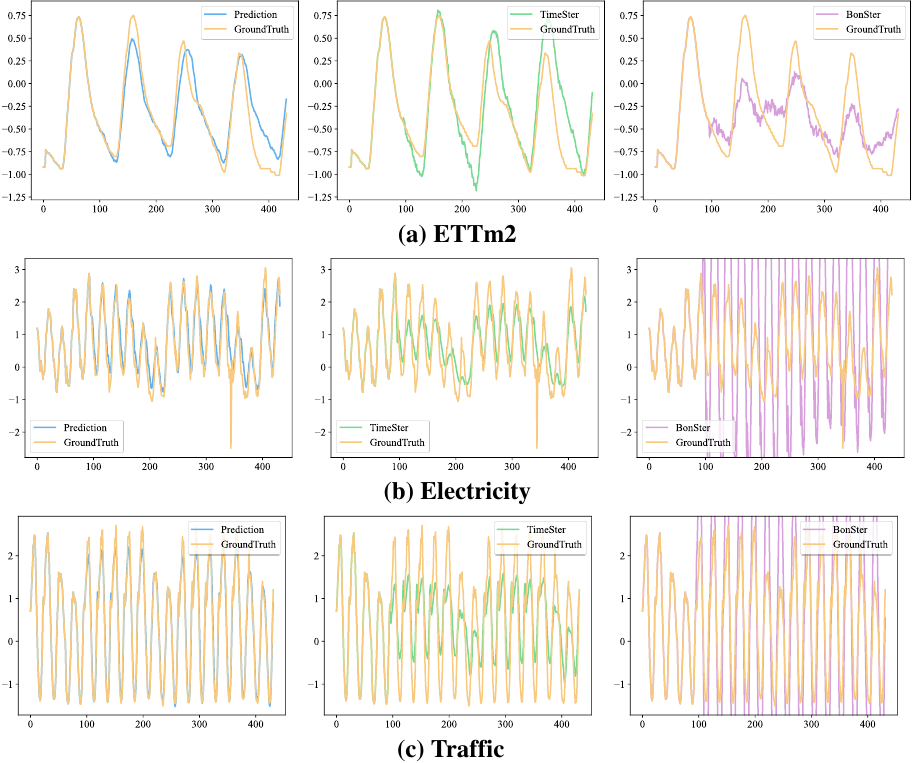}
  \caption{Prediction results of three datasets. $L$ is 96 and $T$ is 336. From left to right, the result of TimeLinear, the result of TimeSter, and the result of BonSter, respectively.}
  \label{fig:vs}
\end{figure}
Generally, a hidden layer that is too large not only reduces model efficiency but also leads to a certain degree of performance degradation, as shown in the results of the Traffic dataset.
On the other hand, a hidden layer size that is too small will lead to insufficient modeling capabilities of the encoder.
Therefore, this is a parameter that needs to be well controlled.

\subsection{Prediction Results Visualization}
\label{sec:visualization}
In this section, we visualize the forecasting results of some datasets under TimeLinear, TimeSter, and BonSter.
As shown in Figure~\ref{fig:vs}, we can see that the TimeSter, which forecasts with time-related features, captures the periodicity and seasonality associated with time stamps well.
In contrast, the BonSter, predicting with historical multivariate time series observations, tends to make predictions that are trend-based but have large deviations from the ground truth due to the influence of historical information.
Although the results of individual predictions are not satisfactory, the combination of the TimeSter and BonSter has a complementary effect, resulting in better prediction accuracy.
The above results not only reflect the importance of time-related features for more accurate long-term predictions but also verify the effectiveness and rationality of our model and pipeline.

\balance
\section{Discussion and Future Work}
\label{sec:conclu}
This paper explores the potential of time-related features in enhancing multivariate long-term time series forecasting.
Methodologically, we propose Time Stamp Forecaster (TimeSter), a plug-and-play module for time series prediction using time-related features.
Extensive experiments demonstrate the versatility of TimeSter on different architectures.
Moreover, its combination with a simple linear layer, named TimeLinear, achieves better performance than state-of-the-art models on multiple datasets while maintaining efficiency.
In addition, we also study the existing shortcomings of our method, including the lack of modeling the relationship between different variables and the decrease in gain brought by time-related features after the historical window is increased.
In the future, we hope to address these deficiencies and explore the possibilities of time-related features in various time series tasks.

\clearpage

\bibliographystyle{ACM-Reference-Format}
\balance
\bibliography{sample}


\begin{thebibliography}{47}


\ifx \showCODEN    \undefined \def \showCODEN     #1{\unskip}     \fi
\ifx \showDOI      \undefined \def \showDOI       #1{#1}\fi
\ifx \showISBNx    \undefined \def \showISBNx     #1{\unskip}     \fi
\ifx \showISBNxiii \undefined \def \showISBNxiii  #1{\unskip}     \fi
\ifx \showISSN     \undefined \def \showISSN      #1{\unskip}     \fi
\ifx \showLCCN     \undefined \def \showLCCN      #1{\unskip}     \fi
\ifx \shownote     \undefined \def \shownote      #1{#1}          \fi
\ifx \showarticletitle \undefined \def \showarticletitle #1{#1}   \fi
\ifx \showURL      \undefined \def \showURL       {\relax}        \fi
\providecommand\bibfield[2]{#2}
\providecommand\bibinfo[2]{#2}
\providecommand\natexlab[1]{#1}
\providecommand\showeprint[2][]{arXiv:#2}

\bibitem[\protect\citeauthoryear{Ansari, Stella, Turkmen, Zhang, Mercado, Shen, Shchur, Rangapuram, Arango, Kapoor, Zschiegner, Maddix, Wang, Mahoney, Torkkola, Wilson, Bohlke-Schneider, and Wang}{Ansari et~al\mbox{.}}{2024}]%
        {ansari2024chronoslearninglanguagetime}
\bibfield{author}{\bibinfo{person}{Abdul~Fatir Ansari}, \bibinfo{person}{Lorenzo Stella}, \bibinfo{person}{Caner Turkmen}, \bibinfo{person}{Xiyuan Zhang}, \bibinfo{person}{Pedro Mercado}, \bibinfo{person}{Huibin Shen}, \bibinfo{person}{Oleksandr Shchur}, \bibinfo{person}{Syama~Sundar Rangapuram}, \bibinfo{person}{Sebastian~Pineda Arango}, \bibinfo{person}{Shubham Kapoor}, \bibinfo{person}{Jasper Zschiegner}, \bibinfo{person}{Danielle~C. Maddix}, \bibinfo{person}{Hao Wang}, \bibinfo{person}{Michael~W. Mahoney}, \bibinfo{person}{Kari Torkkola}, \bibinfo{person}{Andrew~Gordon Wilson}, \bibinfo{person}{Michael Bohlke-Schneider}, {and} \bibinfo{person}{Yuyang Wang}.} \bibinfo{year}{2024}\natexlab{}.
\newblock \bibinfo{title}{Chronos: Learning the Language of Time Series}.
\newblock
\newblock
\showeprint[arxiv]{2403.07815}~[cs.LG]


\bibitem[\protect\citeauthoryear{Bansal, Deshpande, and Sarawagi}{Bansal et~al\mbox{.}}{2021}]%
        {Bansal2021}
\bibfield{author}{\bibinfo{person}{Parikshit Bansal}, \bibinfo{person}{Prathamesh Deshpande}, {and} \bibinfo{person}{Sunita Sarawagi}.} \bibinfo{year}{2021}\natexlab{}.
\newblock \showarticletitle{Missing Value Imputation on Multidimensional Time Series}.
\newblock \bibinfo{journal}{\emph{Proceedings of the VLDB Endowment}} \bibinfo{volume}{14}, \bibinfo{number}{11} (\bibinfo{year}{2021}), \bibinfo{pages}{2533–2545}.
\newblock
\showISSN{2150-8097}


\bibitem[\protect\citeauthoryear{B\"{o}se, Flunkert, Gasthaus, Januschowski, Lange, Salinas, Schelter, Seeger, and Wang}{B\"{o}se et~al\mbox{.}}{2017}]%
        {bose2017}
\bibfield{author}{\bibinfo{person}{Joos-Hendrik B\"{o}se}, \bibinfo{person}{Valentin Flunkert}, \bibinfo{person}{Jan Gasthaus}, \bibinfo{person}{Tim Januschowski}, \bibinfo{person}{Dustin Lange}, \bibinfo{person}{David Salinas}, \bibinfo{person}{Sebastian Schelter}, \bibinfo{person}{Matthias Seeger}, {and} \bibinfo{person}{Yuyang Wang}.} \bibinfo{year}{2017}\natexlab{}.
\newblock \showarticletitle{Probabilistic Demand Forecasting at Scale}.
\newblock \bibinfo{journal}{\emph{Proceedings of the VLDB Endowment}} \bibinfo{volume}{10}, \bibinfo{number}{12} (\bibinfo{year}{2017}), \bibinfo{pages}{1694–1705}.
\newblock
\showISSN{2150-8097}


\bibitem[\protect\citeauthoryear{Box and Jenkins}{Box and Jenkins}{1968}]%
        {box1968some}
\bibfield{author}{\bibinfo{person}{George~EP Box} {and} \bibinfo{person}{Gwilym~M Jenkins}.} \bibinfo{year}{1968}\natexlab{}.
\newblock \showarticletitle{Some Recent Advances in Forecasting and Control}.
\newblock \bibinfo{journal}{\emph{Journal of the Royal Statistical Society. Series C (Applied Statistics)}} \bibinfo{volume}{17}, \bibinfo{number}{2} (\bibinfo{year}{1968}), \bibinfo{pages}{91--109}.
\newblock


\bibitem[\protect\citeauthoryear{Chen, Long, Shen, Jiang, and Zhang}{Chen et~al\mbox{.}}{2024}]%
        {chen2024federatedpromptlearningweather}
\bibfield{author}{\bibinfo{person}{Shengchao Chen}, \bibinfo{person}{Guodong Long}, \bibinfo{person}{Tao Shen}, \bibinfo{person}{Jing Jiang}, {and} \bibinfo{person}{Chengqi Zhang}.} \bibinfo{year}{2024}\natexlab{}.
\newblock \bibinfo{title}{Federated Prompt Learning for Weather Foundation Models on Devices}.
\newblock
\newblock
\showeprint[arxiv]{2305.14244}~[cs.LG]


\bibitem[\protect\citeauthoryear{Cui, Zheng, Cui, Xie, Deng, Huang, and Zhou}{Cui et~al\mbox{.}}{2021}]%
        {cui2021}
\bibfield{author}{\bibinfo{person}{Yue Cui}, \bibinfo{person}{Kai Zheng}, \bibinfo{person}{Dingshan Cui}, \bibinfo{person}{Jiandong Xie}, \bibinfo{person}{Liwei Deng}, \bibinfo{person}{Feiteng Huang}, {and} \bibinfo{person}{Xiaofang Zhou}.} \bibinfo{year}{2021}\natexlab{}.
\newblock \showarticletitle{METRO: A Generic Graph Neural Network Framework for Multivariate Time Series Forecasting}.
\newblock \bibinfo{journal}{\emph{Proceedings of the VLDB Endowment}} \bibinfo{volume}{15}, \bibinfo{number}{2} (\bibinfo{year}{2021}), \bibinfo{pages}{224–236}.
\newblock
\showISSN{2150-8097}


\bibitem[\protect\citeauthoryear{Das, Kong, Leach, Mathur, Sen, and Yu}{Das et~al\mbox{.}}{2024}]%
        {das2023long}
\bibfield{author}{\bibinfo{person}{Abhimanyu Das}, \bibinfo{person}{Weihao Kong}, \bibinfo{person}{Andrew Leach}, \bibinfo{person}{Shaan Mathur}, \bibinfo{person}{Rajat Sen}, {and} \bibinfo{person}{Rose Yu}.} \bibinfo{year}{2024}\natexlab{}.
\newblock \bibinfo{title}{Long-term Forecasting with TiDE: Time-series Dense Encoder}.
\newblock
\newblock
\showeprint[arxiv]{2304.08424}~[stat.ML]


\bibitem[\protect\citeauthoryear{Fang, Pan, Chen, Du, and Gao}{Fang et~al\mbox{.}}{2021}]%
        {fang2021}
\bibfield{author}{\bibinfo{person}{Ziquan Fang}, \bibinfo{person}{Lu Pan}, \bibinfo{person}{Lu Chen}, \bibinfo{person}{Yuntao Du}, {and} \bibinfo{person}{Yunjun Gao}.} \bibinfo{year}{2021}\natexlab{}.
\newblock \showarticletitle{MDTP: A Multi-source Deep Trafic Prediction Framework over Spatio-Temporal Trajectory Data}.
\newblock \bibinfo{journal}{\emph{Proceedings of the VLDB Endowment}} \bibinfo{volume}{14}, \bibinfo{number}{8} (\bibinfo{year}{2021}), \bibinfo{pages}{1289–1297}.
\newblock
\showISSN{2150-8097}


\bibitem[\protect\citeauthoryear{Goswami, Szafer, Choudhry, Cai, Li, and Dubrawski}{Goswami et~al\mbox{.}}{2024}]%
        {goswami2024moment}
\bibfield{author}{\bibinfo{person}{Mononito Goswami}, \bibinfo{person}{Konrad Szafer}, \bibinfo{person}{Arjun Choudhry}, \bibinfo{person}{Yifu Cai}, \bibinfo{person}{Shuo Li}, {and} \bibinfo{person}{Artur Dubrawski}.} \bibinfo{year}{2024}\natexlab{}.
\newblock \showarticletitle{MOMENT: A Family of Open Time-series Foundation Models}. In \bibinfo{booktitle}{\emph{International Conference on Machine Learning}}.
\newblock


\bibitem[\protect\citeauthoryear{Han, Chen, Ye, and Zhan}{Han et~al\mbox{.}}{2024a}]%
        {han2024softs}
\bibfield{author}{\bibinfo{person}{Lu Han}, \bibinfo{person}{Xu-Yang Chen}, \bibinfo{person}{Han-Jia Ye}, {and} \bibinfo{person}{De-Chuan Zhan}.} \bibinfo{year}{2024}\natexlab{a}.
\newblock \showarticletitle{SOFTS: Efficient Multivariate Time Series Forecasting with Series-Core Fusion}. In \bibinfo{booktitle}{\emph{Advances in Neural Information Processing Systems}}.
\newblock


\bibitem[\protect\citeauthoryear{Han, Ye, and Zhan}{Han et~al\mbox{.}}{2024b}]%
        {han2023capacity}
\bibfield{author}{\bibinfo{person}{Lu Han}, \bibinfo{person}{Han-Jia Ye}, {and} \bibinfo{person}{De-Chuan Zhan}.} \bibinfo{year}{2024}\natexlab{b}.
\newblock \showarticletitle{The Capacity and Robustness Trade-off: Revisiting the Channel Independent Strategy for Multivariate Time Series Forecasting}.
\newblock \bibinfo{journal}{\emph{IEEE Transactions on Knowledge and Data Engineering}} \bibinfo{volume}{36}, \bibinfo{number}{11} (\bibinfo{year}{2024}), \bibinfo{pages}{7129–7142}.
\newblock


\bibitem[\protect\citeauthoryear{Jin, Wang, Ma, Chu, Zhang, Shi, Chen, Liang, Li, Pan, et~al\mbox{.}}{Jin et~al\mbox{.}}{2024}]%
        {jin2023time}
\bibfield{author}{\bibinfo{person}{Ming Jin}, \bibinfo{person}{Shiyu Wang}, \bibinfo{person}{Lintao Ma}, \bibinfo{person}{Zhixuan Chu}, \bibinfo{person}{James~Y Zhang}, \bibinfo{person}{Xiaoming Shi}, \bibinfo{person}{Pin-Yu Chen}, \bibinfo{person}{Yuxuan Liang}, \bibinfo{person}{Yuan-Fang Li}, \bibinfo{person}{Shirui Pan}, {et~al\mbox{.}}} \bibinfo{year}{2024}\natexlab{}.
\newblock \showarticletitle{Time-LLM: Time Series Forecasting by Reprogramming Large Language Models}. In \bibinfo{booktitle}{\emph{International Conference on Learning Representations}}.
\newblock


\bibitem[\protect\citeauthoryear{Khayati, Lerner, Tymchenko, and Cudr\'{e}-Mauroux}{Khayati et~al\mbox{.}}{2020}]%
        {khayati2020}
\bibfield{author}{\bibinfo{person}{Mourad Khayati}, \bibinfo{person}{Alberto Lerner}, \bibinfo{person}{Zakhar Tymchenko}, {and} \bibinfo{person}{Philippe Cudr\'{e}-Mauroux}.} \bibinfo{year}{2020}\natexlab{}.
\newblock \showarticletitle{Mind the Gap: An Experimental Evaluation of Imputation of Missing Values Techniques in Time Series}.
\newblock \bibinfo{journal}{\emph{Proceedings of the VLDB Endowment}} \bibinfo{volume}{13}, \bibinfo{number}{5} (\bibinfo{year}{2020}), \bibinfo{pages}{768–782}.
\newblock
\showISSN{2150-8097}


\bibitem[\protect\citeauthoryear{Kim, Kim, Tae, Park, Choi, and Choo}{Kim et~al\mbox{.}}{2021}]%
        {kim2021reversible}
\bibfield{author}{\bibinfo{person}{Taesung Kim}, \bibinfo{person}{Jinhee Kim}, \bibinfo{person}{Yunwon Tae}, \bibinfo{person}{Cheonbok Park}, \bibinfo{person}{Jang-Ho Choi}, {and} \bibinfo{person}{Jaegul Choo}.} \bibinfo{year}{2021}\natexlab{}.
\newblock \showarticletitle{Reversible Instance Normalization for Accurate Time-Series Forecasting against Distribution Shift}. In \bibinfo{booktitle}{\emph{International Conference on Learning Representations}}.
\newblock


\bibitem[\protect\citeauthoryear{Kingma and Ba}{Kingma and Ba}{2017}]%
        {kingma2014adam}
\bibfield{author}{\bibinfo{person}{Diederik~P. Kingma} {and} \bibinfo{person}{Jimmy Ba}.} \bibinfo{year}{2017}\natexlab{}.
\newblock \bibinfo{title}{Adam: A Method for Stochastic Optimization}.
\newblock
\newblock
\showeprint[arxiv]{1412.6980}~[cs.LG]


\bibitem[\protect\citeauthoryear{Kingma and Welling}{Kingma and Welling}{2014}]%
        {kingma2013auto}
\bibfield{author}{\bibinfo{person}{Diederik~P Kingma} {and} \bibinfo{person}{Max Welling}.} \bibinfo{year}{2014}\natexlab{}.
\newblock \showarticletitle{Auto-Encoding Variational Bayes}. In \bibinfo{booktitle}{\emph{International Conference on Learning Representations}}.
\newblock


\bibitem[\protect\citeauthoryear{Lai, Chang, Yang, and Liu}{Lai et~al\mbox{.}}{2018}]%
        {lai2018modeling}
\bibfield{author}{\bibinfo{person}{Guokun Lai}, \bibinfo{person}{Wei-Cheng Chang}, \bibinfo{person}{Yiming Yang}, {and} \bibinfo{person}{Hanxiao Liu}.} \bibinfo{year}{2018}\natexlab{}.
\newblock \showarticletitle{Modeling Long- and Short-Term Temporal Patterns with Deep Neural Networks}. In \bibinfo{booktitle}{\emph{International ACM SIGIR Conference on Research and Development in Information Retrieval}}.
\newblock


\bibitem[\protect\citeauthoryear{Li, Qi, Li, and Xu}{Li et~al\mbox{.}}{2023}]%
        {li2023revisiting}
\bibfield{author}{\bibinfo{person}{Zhe Li}, \bibinfo{person}{Shiyi Qi}, \bibinfo{person}{Yiduo Li}, {and} \bibinfo{person}{Zenglin Xu}.} \bibinfo{year}{2023}\natexlab{}.
\newblock \bibinfo{title}{Revisiting Long-term Time Series Forecasting: An Investigation on Linear Mapping}.
\newblock
\newblock
\showeprint[arxiv]{2305.10721}~[cs.LG]


\bibitem[\protect\citeauthoryear{Lin, Lin, Hu, Wu, Mo, and Zhong}{Lin et~al\mbox{.}}{2024}]%
        {lin2024cyclenet}
\bibfield{author}{\bibinfo{person}{Shengsheng Lin}, \bibinfo{person}{Weiwei Lin}, \bibinfo{person}{Xinyi Hu}, \bibinfo{person}{Wentai Wu}, \bibinfo{person}{Ruichao Mo}, {and} \bibinfo{person}{Haocheng Zhong}.} \bibinfo{year}{2024}\natexlab{}.
\newblock \showarticletitle{CycleNet: Enhancing Time Series Forecasting through Modeling Periodic Patterns}. In \bibinfo{booktitle}{\emph{Advances in Neural Information Processing Systems}}.
\newblock


\bibitem[\protect\citeauthoryear{Liu, Hu, Zhang, Wu, Wang, Ma, and Long}{Liu et~al\mbox{.}}{2024a}]%
        {liu2024itransformer}
\bibfield{author}{\bibinfo{person}{Yong Liu}, \bibinfo{person}{Tengge Hu}, \bibinfo{person}{Haoran Zhang}, \bibinfo{person}{Haixu Wu}, \bibinfo{person}{Shiyu Wang}, \bibinfo{person}{Lintao Ma}, {and} \bibinfo{person}{Mingsheng Long}.} \bibinfo{year}{2024}\natexlab{a}.
\newblock \showarticletitle{iTransformer: Inverted Transformers Are Effective for Time Series Forecasting}. In \bibinfo{booktitle}{\emph{International Conference on Learning Representations}}.
\newblock


\bibitem[\protect\citeauthoryear{Liu, Qin, Huang, Wang, and Long}{Liu et~al\mbox{.}}{2024b}]%
        {liu2024autotimes}
\bibfield{author}{\bibinfo{person}{Yong Liu}, \bibinfo{person}{Guo Qin}, \bibinfo{person}{Xiangdong Huang}, \bibinfo{person}{Jianmin Wang}, {and} \bibinfo{person}{Mingsheng Long}.} \bibinfo{year}{2024}\natexlab{b}.
\newblock \showarticletitle{AutoTimes: Autoregressive Time Series Forecasters via Large Language Models}. In \bibinfo{booktitle}{\emph{Advances in Neural Information Processing Systems}}.
\newblock


\bibitem[\protect\citeauthoryear{Luo and Wang}{Luo and Wang}{2024}]%
        {luo2024moderntcn}
\bibfield{author}{\bibinfo{person}{Donghao Luo} {and} \bibinfo{person}{Xue Wang}.} \bibinfo{year}{2024}\natexlab{}.
\newblock \showarticletitle{ModernTCN: A Modern Pure Convolution Structure for General Time Series Analysis}. In \bibinfo{booktitle}{\emph{International Conference on Learning Representations}}.
\newblock


\bibitem[\protect\citeauthoryear{Lv, Duan, Kang, Li, and Wang}{Lv et~al\mbox{.}}{2015}]%
        {lv2014traffic}
\bibfield{author}{\bibinfo{person}{Yisheng Lv}, \bibinfo{person}{Yanjie Duan}, \bibinfo{person}{Wenwen Kang}, \bibinfo{person}{Zhengxi Li}, {and} \bibinfo{person}{Fei-Yue Wang}.} \bibinfo{year}{2015}\natexlab{}.
\newblock \showarticletitle{Traffic Flow Prediction With Big Data: A Deep Learning Approach}.
\newblock \bibinfo{journal}{\emph{IEEE Transactions on Intelligent Transportation Systems}} \bibinfo{volume}{16}, \bibinfo{number}{2} (\bibinfo{year}{2015}), \bibinfo{pages}{865--873}.
\newblock


\bibitem[\protect\citeauthoryear{Madsen}{Madsen}{2007}]%
        {madsen2007time}
\bibfield{author}{\bibinfo{person}{Henrik Madsen}.} \bibinfo{year}{2007}\natexlab{}.
\newblock \bibinfo{booktitle}{\emph{Time Series Analysis}}.
\newblock \bibinfo{publisher}{Chapman and Hall/CRC}.
\newblock


\bibitem[\protect\citeauthoryear{Narasimhan, Agarwal, Akcin, Sanghavi, and Chinchali}{Narasimhan et~al\mbox{.}}{2024}]%
        {narasimhan2024time}
\bibfield{author}{\bibinfo{person}{Sai~Shankar Narasimhan}, \bibinfo{person}{Shubhankar Agarwal}, \bibinfo{person}{Oguzhan Akcin}, \bibinfo{person}{Sujay Sanghavi}, {and} \bibinfo{person}{Sandeep Chinchali}.} \bibinfo{year}{2024}\natexlab{}.
\newblock \showarticletitle{Time Weaver: A Conditional Time Series Generation Model}. In \bibinfo{booktitle}{\emph{International Conference on Machine Learning}}.
\newblock


\bibitem[\protect\citeauthoryear{Nie, H.~Nguyen, Sinthong, and Kalagnanam}{Nie et~al\mbox{.}}{2023}]%
        {Yuqietal-2023-PatchTST}
\bibfield{author}{\bibinfo{person}{Yuqi Nie}, \bibinfo{person}{Nam H.~Nguyen}, \bibinfo{person}{Phanwadee Sinthong}, {and} \bibinfo{person}{Jayant Kalagnanam}.} \bibinfo{year}{2023}\natexlab{}.
\newblock \showarticletitle{A Time Series is Worth 64 Words: Long-term Forecasting with Transformers}. In \bibinfo{booktitle}{\emph{International Conference on Learning Representations}}.
\newblock


\bibitem[\protect\citeauthoryear{Paparrizos, Kang, Boniol, Tsay, Palpanas, and Franklin}{Paparrizos et~al\mbox{.}}{2022}]%
        {paparrizos2022}
\bibfield{author}{\bibinfo{person}{John Paparrizos}, \bibinfo{person}{Yuhao Kang}, \bibinfo{person}{Paul Boniol}, \bibinfo{person}{Ruey~S. Tsay}, \bibinfo{person}{Themis Palpanas}, {and} \bibinfo{person}{Michael~J. Franklin}.} \bibinfo{year}{2022}\natexlab{}.
\newblock \showarticletitle{TSB-UAD: An End-to-End Benchmark Suite for Univariate Time-Series Anomaly Detection}.
\newblock \bibinfo{journal}{\emph{Proceedings of the VLDB Endowment}} \bibinfo{volume}{15}, \bibinfo{number}{8} (\bibinfo{year}{2022}), \bibinfo{pages}{1697–1711}.
\newblock
\showISSN{2150-8097}


\bibitem[\protect\citeauthoryear{Paszke, Gross, Massa, Lerer, Bradbury, Chanan, Killeen, Lin, Gimelshein, Antiga, et~al\mbox{.}}{Paszke et~al\mbox{.}}{2019}]%
        {paszke2019pytorch}
\bibfield{author}{\bibinfo{person}{Adam Paszke}, \bibinfo{person}{Sam Gross}, \bibinfo{person}{Francisco Massa}, \bibinfo{person}{Adam Lerer}, \bibinfo{person}{James Bradbury}, \bibinfo{person}{Gregory Chanan}, \bibinfo{person}{Trevor Killeen}, \bibinfo{person}{Zeming Lin}, \bibinfo{person}{Natalia Gimelshein}, \bibinfo{person}{Luca Antiga}, {et~al\mbox{.}}} \bibinfo{year}{2019}\natexlab{}.
\newblock \showarticletitle{PyTorch: An Imperative Style, High-Performance Deep Learning Library}. In \bibinfo{booktitle}{\emph{Advances in Neural Information Processing Systems}}.
\newblock


\bibitem[\protect\citeauthoryear{Pereira, Salazar, and Vergara}{Pereira et~al\mbox{.}}{2023}]%
        {pereira2023comparative}
\bibfield{author}{\bibinfo{person}{Luis~Manuel Pereira}, \bibinfo{person}{Addisson Salazar}, {and} \bibinfo{person}{Luis Vergara}.} \bibinfo{year}{2023}\natexlab{}.
\newblock \showarticletitle{A Comparative Analysis of Early and Late Fusion for the Multimodal Two-Class Problem}.
\newblock \bibinfo{journal}{\emph{IEEE Access}}  \bibinfo{volume}{11} (\bibinfo{year}{2023}), \bibinfo{pages}{84283--84300}.
\newblock


\bibitem[\protect\citeauthoryear{Qiu, Hu, Zhou, Wu, Du, Zhang, Guo, Zhou, Jensen, Sheng, and Yang}{Qiu et~al\mbox{.}}{2024}]%
        {qiu2024tfb}
\bibfield{author}{\bibinfo{person}{Xiangfei Qiu}, \bibinfo{person}{Jilin Hu}, \bibinfo{person}{Lekui Zhou}, \bibinfo{person}{Xingjian Wu}, \bibinfo{person}{Junyang Du}, \bibinfo{person}{Buang Zhang}, \bibinfo{person}{Chenjuan Guo}, \bibinfo{person}{Aoying Zhou}, \bibinfo{person}{Christian~S. Jensen}, \bibinfo{person}{Zhenli Sheng}, {and} \bibinfo{person}{Bin Yang}.} \bibinfo{year}{2024}\natexlab{}.
\newblock \showarticletitle{TFB: Towards Comprehensive and Fair Benchmarking of Time Series Forecasting Methods}.
\newblock \bibinfo{journal}{\emph{Proceedings of the VLDB Endowment}} \bibinfo{volume}{17}, \bibinfo{number}{9} (\bibinfo{year}{2024}), \bibinfo{pages}{2363--2377}.
\newblock


\bibitem[\protect\citeauthoryear{Schmidl, Wenig, and Papenbrock}{Schmidl et~al\mbox{.}}{2022}]%
        {SchmidlEtAl2022Anomaly}
\bibfield{author}{\bibinfo{person}{Sebastian Schmidl}, \bibinfo{person}{Phillip Wenig}, {and} \bibinfo{person}{Thorsten Papenbrock}.} \bibinfo{year}{2022}\natexlab{}.
\newblock \showarticletitle{Anomaly Detection in Time Series: A Comprehensive Evaluation}.
\newblock \bibinfo{journal}{\emph{Proceedings of the VLDB Endowment}} \bibinfo{volume}{15}, \bibinfo{number}{9} (\bibinfo{year}{2022}), \bibinfo{pages}{1779--1797}.
\newblock


\bibitem[\protect\citeauthoryear{Shi}{Shi}{2024}]%
        {shi2024mambastock}
\bibfield{author}{\bibinfo{person}{Zhuangwei Shi}.} \bibinfo{year}{2024}\natexlab{}.
\newblock \bibinfo{title}{MambaStock: Selective state space model for stock prediction}.
\newblock
\newblock
\showeprint[arxiv]{2402.18959}~[cs.CE]


\bibitem[\protect\citeauthoryear{Tran, Mun, Lim, Yamato, Huh, and Shahabi}{Tran et~al\mbox{.}}{2020}]%
        {tran2020}
\bibfield{author}{\bibinfo{person}{Luan Tran}, \bibinfo{person}{Min~Y. Mun}, \bibinfo{person}{Matthew Lim}, \bibinfo{person}{Jonah Yamato}, \bibinfo{person}{Nathan Huh}, {and} \bibinfo{person}{Cyrus Shahabi}.} \bibinfo{year}{2020}\natexlab{}.
\newblock \showarticletitle{DeepTRANS: A Deep Learning System for Public Bus Travel Time Estimation using Traffic Forecasting}.
\newblock \bibinfo{journal}{\emph{Proceedings of the VLDB Endowment}} \bibinfo{volume}{13}, \bibinfo{number}{12} (\bibinfo{year}{2020}), \bibinfo{pages}{2957–2960}.
\newblock
\showISSN{2150-8097}


\bibitem[\protect\citeauthoryear{Wang, Qi, Wang, Sun, Zhuang, Wu, and Liao}{Wang et~al\mbox{.}}{2024b}]%
        {wang2024rethinking}
\bibfield{author}{\bibinfo{person}{Chengsen Wang}, \bibinfo{person}{Qi Qi}, \bibinfo{person}{Jingyu Wang}, \bibinfo{person}{Haifeng Sun}, \bibinfo{person}{Zirui Zhuang}, \bibinfo{person}{Jinming Wu}, {and} \bibinfo{person}{Jianxin Liao}.} \bibinfo{year}{2024}\natexlab{b}.
\newblock \showarticletitle{Rethinking the Power of Timestamps for Robust Time Series Forecasting: A Global-Local Fusion Perspective}. In \bibinfo{booktitle}{\emph{Advances in Neural Information Processing Systems}}.
\newblock


\bibitem[\protect\citeauthoryear{Wang, Peng, Huang, Wang, Chen, and Xiao}{Wang et~al\mbox{.}}{2023}]%
        {wang2022micn}
\bibfield{author}{\bibinfo{person}{Huiqiang Wang}, \bibinfo{person}{Jian Peng}, \bibinfo{person}{Feihu Huang}, \bibinfo{person}{Jince Wang}, \bibinfo{person}{Junhui Chen}, {and} \bibinfo{person}{Yifei Xiao}.} \bibinfo{year}{2023}\natexlab{}.
\newblock \showarticletitle{MICN: Multi-scale Local and Global Context Modeling for Long-term Series Forecasting}. In \bibinfo{booktitle}{\emph{International Conference on Learning Representations}}.
\newblock


\bibitem[\protect\citeauthoryear{Wang, Wu, Shi, Hu, Luo, Ma, Zhang, and ZHOU}{Wang et~al\mbox{.}}{2024c}]%
        {wang2023timemixer}
\bibfield{author}{\bibinfo{person}{Shiyu Wang}, \bibinfo{person}{Haixu Wu}, \bibinfo{person}{Xiaoming Shi}, \bibinfo{person}{Tengge Hu}, \bibinfo{person}{Huakun Luo}, \bibinfo{person}{Lintao Ma}, \bibinfo{person}{James~Y Zhang}, {and} \bibinfo{person}{JUN ZHOU}.} \bibinfo{year}{2024}\natexlab{c}.
\newblock \showarticletitle{TimeMixer: Decomposable Multiscale Mixing for Time Series Forecasting}. In \bibinfo{booktitle}{\emph{International Conference on Learning Representations}}.
\newblock


\bibitem[\protect\citeauthoryear{Wang, Feng, Qiu, Gu, and Zhao}{Wang et~al\mbox{.}}{2024a}]%
        {wang2024news}
\bibfield{author}{\bibinfo{person}{Xinlei Wang}, \bibinfo{person}{Maike Feng}, \bibinfo{person}{Jing Qiu}, \bibinfo{person}{Jinjin Gu}, {and} \bibinfo{person}{Junhua Zhao}.} \bibinfo{year}{2024}\natexlab{a}.
\newblock \showarticletitle{From News to Forecast: Integrating Event Analysis in LLM-Based Time Series Forecasting with Reflection}. In \bibinfo{booktitle}{\emph{Advances in Neural Information Processing Systems}}.
\newblock


\bibitem[\protect\citeauthoryear{Wenig, Schmidl, and Papenbrock}{Wenig et~al\mbox{.}}{2022}]%
        {WenigEtAl2022TimeEval}
\bibfield{author}{\bibinfo{person}{Phillip Wenig}, \bibinfo{person}{Sebastian Schmidl}, {and} \bibinfo{person}{Thorsten Papenbrock}.} \bibinfo{year}{2022}\natexlab{}.
\newblock \showarticletitle{TimeEval: A Benchmarking Toolkit for Time Series Anomaly Detection Algorithms}.
\newblock \bibinfo{journal}{\emph{Proceedings of the VLDB Endowment}} \bibinfo{volume}{15}, \bibinfo{number}{12} (\bibinfo{year}{2022}), \bibinfo{pages}{3678--3681}.
\newblock


\bibitem[\protect\citeauthoryear{Wu, Hu, Liu, Zhou, Wang, and Long}{Wu et~al\mbox{.}}{2023}]%
        {wu2023timesnet}
\bibfield{author}{\bibinfo{person}{Haixu Wu}, \bibinfo{person}{Tengge Hu}, \bibinfo{person}{Yong Liu}, \bibinfo{person}{Hang Zhou}, \bibinfo{person}{Jianmin Wang}, {and} \bibinfo{person}{Mingsheng Long}.} \bibinfo{year}{2023}\natexlab{}.
\newblock \showarticletitle{TimesNet: Temporal 2D-Variation Modeling for General Time Series Analysis}. In \bibinfo{booktitle}{\emph{International Conference on Learning Representations}}.
\newblock


\bibitem[\protect\citeauthoryear{Wu, Xu, Wang, and Long}{Wu et~al\mbox{.}}{2021}]%
        {wu2021autoformer}
\bibfield{author}{\bibinfo{person}{Haixu Wu}, \bibinfo{person}{Jiehui Xu}, \bibinfo{person}{Jianmin Wang}, {and} \bibinfo{person}{Mingsheng Long}.} \bibinfo{year}{2021}\natexlab{}.
\newblock \showarticletitle{Autoformer: Decomposition Transformers with {Auto-Correlation} for Long-Term Series Forecasting}. In \bibinfo{booktitle}{\emph{Advances in Neural Information Processing Systems}}.
\newblock


\bibitem[\protect\citeauthoryear{Xu and Cohen}{Xu and Cohen}{2018}]%
        {xu2018stock}
\bibfield{author}{\bibinfo{person}{Yumo Xu} {and} \bibinfo{person}{Shay~B Cohen}.} \bibinfo{year}{2018}\natexlab{}.
\newblock \showarticletitle{Stock Movement Prediction from Tweets and Historical Prices}. In \bibinfo{booktitle}{\emph{Proceedings of the 56th Annual Meeting of the Association for Computational Linguistics (Volume 1: Long Papers)}}.
\newblock


\bibitem[\protect\citeauthoryear{Xu, Zeng, and Xu}{Xu et~al\mbox{.}}{2024}]%
        {xu2023fits}
\bibfield{author}{\bibinfo{person}{Zhijian Xu}, \bibinfo{person}{Ailing Zeng}, {and} \bibinfo{person}{Qiang Xu}.} \bibinfo{year}{2024}\natexlab{}.
\newblock \showarticletitle{FITS: Modeling Time Series with $10k$ Parameters}. In \bibinfo{booktitle}{\emph{International Conference on Learning Representations}}.
\newblock


\bibitem[\protect\citeauthoryear{Zeng, Chen, Zhang, and Xu}{Zeng et~al\mbox{.}}{2023}]%
        {zeng2023transformers}
\bibfield{author}{\bibinfo{person}{Ailing Zeng}, \bibinfo{person}{Muxi Chen}, \bibinfo{person}{Lei Zhang}, {and} \bibinfo{person}{Qiang Xu}.} \bibinfo{year}{2023}\natexlab{}.
\newblock \showarticletitle{Are Transformers Effective for Time Series Forecasting?}. In \bibinfo{booktitle}{\emph{Proceedings of the AAAI Conference on Artificial Intelligence}}.
\newblock


\bibitem[\protect\citeauthoryear{Zhang, Long, Chen, Xing, Jin, Jordan, and Wang}{Zhang et~al\mbox{.}}{2023}]%
        {zhang2023skilful}
\bibfield{author}{\bibinfo{person}{Yuchen Zhang}, \bibinfo{person}{Mingsheng Long}, \bibinfo{person}{Kaiyuan Chen}, \bibinfo{person}{Lanxiang Xing}, \bibinfo{person}{Ronghua Jin}, \bibinfo{person}{Michael~I Jordan}, {and} \bibinfo{person}{Jianmin Wang}.} \bibinfo{year}{2023}\natexlab{}.
\newblock \showarticletitle{Skilful Nowcasting of Extreme Precipitation with NowcastNet}.
\newblock \bibinfo{journal}{\emph{Nature}} \bibinfo{volume}{619}, \bibinfo{number}{7970} (\bibinfo{year}{2023}), \bibinfo{pages}{526--532}.
\newblock


\bibitem[\protect\citeauthoryear{Zhang and Yan}{Zhang and Yan}{2023}]%
        {zhang2023crossformer}
\bibfield{author}{\bibinfo{person}{Yunhao Zhang} {and} \bibinfo{person}{Junchi Yan}.} \bibinfo{year}{2023}\natexlab{}.
\newblock \showarticletitle{Crossformer: Transformer Utilizing Cross-Dimension Dependency for Multivariate Time Series Forecasting}. In \bibinfo{booktitle}{\emph{International Conference on Learning Representations}}.
\newblock


\bibitem[\protect\citeauthoryear{Zhong, Song, Zhuo, Li, Liu, and Chan}{Zhong et~al\mbox{.}}{2024}]%
        {zhong2024}
\bibfield{author}{\bibinfo{person}{Shuhan Zhong}, \bibinfo{person}{Sizhe Song}, \bibinfo{person}{Weipeng Zhuo}, \bibinfo{person}{Guanyao Li}, \bibinfo{person}{Yang Liu}, {and} \bibinfo{person}{S.-H.~Gary Chan}.} \bibinfo{year}{2024}\natexlab{}.
\newblock \showarticletitle{A Multi-Scale Decomposition MLP-Mixer for Time Series Analysis}.
\newblock \bibinfo{journal}{\emph{Proceedings of the VLDB Endowment}} \bibinfo{volume}{17}, \bibinfo{number}{7} (\bibinfo{year}{2024}), \bibinfo{pages}{1723–1736}.
\newblock
\showISSN{2150-8097}


\bibitem[\protect\citeauthoryear{Zhou, Zhang, Peng, Zhang, Li, Xiong, and Zhang}{Zhou et~al\mbox{.}}{2021}]%
        {zhou2021informer}
\bibfield{author}{\bibinfo{person}{Haoyi Zhou}, \bibinfo{person}{Shanghang Zhang}, \bibinfo{person}{Jieqi Peng}, \bibinfo{person}{Shuai Zhang}, \bibinfo{person}{Jianxin Li}, \bibinfo{person}{Hui Xiong}, {and} \bibinfo{person}{Wancai Zhang}.} \bibinfo{year}{2021}\natexlab{}.
\newblock \showarticletitle{Informer: Beyond Efficient Transformer for Long Sequence Time-Series Forecasting}. In \bibinfo{booktitle}{\emph{Proceedings of the AAAI Conference on Artificial Intelligence}}.
\newblock


\end{thebibliography}

\end{document}